\documentclass[twocolumn,letter]{IEEEtran}
\usepackage[flushleft]{threeparttable}
\usepackage{cite}
\usepackage{color,soul}
\usepackage{times}
\usepackage{amsmath}
\usepackage{psfrag}
\usepackage{bm}
\usepackage{dblfloatfix}
\usepackage{mathtools}
\usepackage{amsmath}
\usepackage{amsthm}
\usepackage{amssymb}
\usepackage{subfigure}
\usepackage{verbatim}
\usepackage{graphicx}
\usepackage[thinlines,thiklines]{easybmat}
\usepackage{latexsym}
\usepackage[dvipsnames]{xcolor}
\usepackage{cite}
\usepackage{stmaryrd}
\usepackage{cuted}
\usepackage{lipsum}
\usepackage{float}
\usepackage{multirow}
\usepackage{textcomp}
\usepackage[ruled,vlined]{algorithm2e}
\usepackage{bookmark}
\usepackage[most]{tcolorbox}

\newsavebox{\twosubbox}

\tcbset{
    frame code={}
    center title,
    left=0pt,
    right=0pt,
    top=0pt,
    bottom=0pt,
    colback=gray!70,
    colframe=white,
    width=\dimexpr\textwidth\relax,
    enlarge left by=0mm,
    boxsep=5pt,
    arc=0pt,outer arc=0pt,
    }

\newcommand{\bs}[1]{\ensuremath{{\boldsymbol{#1}}}}

\def\diag{\mathop{\rm diag}\nolimits}

\def\diag{\mathop{\rm diag}\nolimits}

\makeatletter
\def\ALG@special@indent{%
	\ifdim\ALG@thistlm=0pt\relax
	\hskip-\leftmargin
	\else
	\hskip\ALG@thistlm
	\fi
}
\newcommand{\Notations}[1]{\item[]\noindent\ALG@special@indent \textbf{Notations:}\ #1}

\makeatletter
\newcommand{\multiline}[1]{%
	\begin{tabularx}{\dimexpr\linewidth-\ALG@thistlm}[t]{@{}X@{}}
		#1
	\end{tabularx}
}
\makeatother

\DeclareFontFamily{OMX}{yhex}{}
\DeclareFontShape{OMX}{yhex}{m}{n}{<->yhcmex10}{}
\DeclareSymbolFont{yhlargesymbols}{OMX}{yhex}{m}{n}
\DeclareMathAccent{\wideparen}{\mathord}{yhlargesymbols}{"F3}

\title{Coordinated Pose Control of Mobile Manipulation with an Unstable Bikebot Platform~\thanks{This work was partially supported by the US National Science Foundation under award CNS-1932370.}}

\author{Feng Han\thanks{F. Han and J. Yi are with the Department of Mechanical and Aerospace Engineering, Rutgers University, Piscataway, NJ 08854 USA (e-mail: fh233@scarletmail.rutgers.edu; jgyi@rutgers.edu).}, Alborz Jelvani\thanks{A. Jelvani is with the  Department of Computer Science, Rutgers University, Piscataway, NJ 08854 USA (e-mail: aj654@scarletmail.rutgers.edu).}, Jingang Yi, and Tao Liu\thanks{T. Liu is with the State Key Lab of Fluid Power and Mechatronic Systems and the School of Mechanical Engineering, Zhejiang University, Hangzhou, Zhejiang 310027 China (email: {liutao@zju.edu.cn}).}}

\begin{document}
\maketitle

\begin{abstract}
Bikebot manipulation has advantages of the single-track robot mobility and manipulation dexterity. We present a coordinated pose control of mobile manipulation with the stationary bikebot. The challenges of the bikebot manipulation include the limited steering balance capability of the unstable bikebot and kinematic redundancy of the manipulator. We first present the steering balance model to analyze and explore the maximum steering capability to balance the stationary platform. A balancing equilibrium manifold is then proposed to describe the necessary condition to fulfill the simultaneous platform balance and posture control of the end-effector. A coordinated planning and control design is presented to determine the balance-prioritized posture control under kinematic and dynamic constraints. Extensive experiments are conducted to demonstrate the mechatronic design for autonomous plant inspection in agricultural applications. The results confirm the feasibility to use the bikebot manipulation for a plant inspection with end-effector position and orientation errors about 5 mm and 0.3 degs, respectively.
\end{abstract}


\section{Introduction}
\label{Section_Introduction}

Mobile manipulation integrates a mobile robot with an onboard multi-link manipulator to expand workspace and improve capability for complex manipulation tasks~\cite{Lu2012surveyarms,Brock2016,SIMETTI2019103287}. Mobile manipulation can be built on wheeled, legged or aerial platforms and the applications include agriculture harvesting~\cite{Arad2020pepper}, mobile cranes~\cite{nguyen2015stability}, underwater archaeology~\cite{stuart2017ocean}, and aerial manipulation~\cite{Kim2016Vision,BONYANKHAMSEH2018221}, etc. The advantages of the mobile manipulation come at the cost of coordinated planning and control~\cite{Patrick2012Reachable}. The coupled dynamics of the mobile platform and manipulator is one of the design challenges~\cite{RAJA2019103245}. Unknown or uncertain robot-environment interactions bring additional complexity for control of mobile manipulation~\cite{nguyen2015stability,Jon2019Control}. For instance, ocean waves and tides cause large dynamic disturbances to the ship-mounted manipulator~\cite{FromICRA2009} and a wheeled/legged mobile robot would fall down when moving on a steep or rocky field~\cite{RAIBERT200810822}. For aerial manipulation, interaction forces generates large disturbances for robot motion and control due to small masses and limited actuation of the quadrotors~\cite{Kim2016Vision,Kim2018Cooperative}.

Coordinated planning and control is critical when the mobile platform is unstable or in complex, dynamic environments. Balance control of unstable platform is among the highest priority tasks for mobile manipulation. In~\cite{Minniti2019Whole-Body}, a model predictive control was presented for collaborative manipulation, balancing and interaction of a ball-based three degree-of-freedom (DOF) manipulator. In~\cite{Nagarajan2009Trajectory},  an unstable mobile manipulation used a single spherical wheel-based ``ballbot'' as the mobile platform and a balance motion control was developed for the underactuated, nonholonomic robot. For kinematic redundant manipulators, task-priority control takes advantages of design space in the null space of the Jacobian matrix. Velocity control was designed through optimization to satisfy the control tasks from the highest to lowest priorities~\cite{Moe2016Tasks,SIMETTI201840Task,Kanoun2011Kinematic}.

\begin{figure*}[htp!]
	\centering
	\subfigure[]{
  \label{Fig_Bike_Photo}
		\includegraphics[width=2.6in]{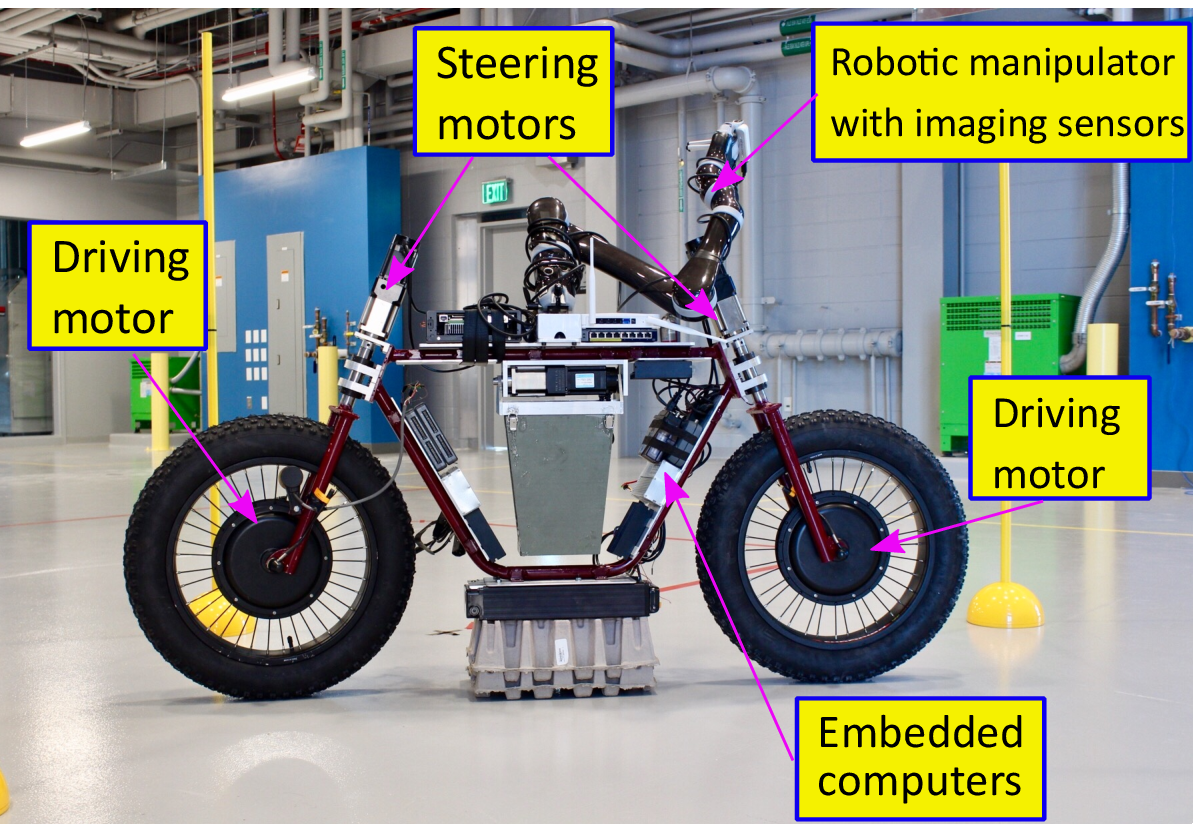}}
	\hspace{3mm}
	\subfigure[]{
  \label{Fig_Schemetics}
		\includegraphics[width=2.55in]{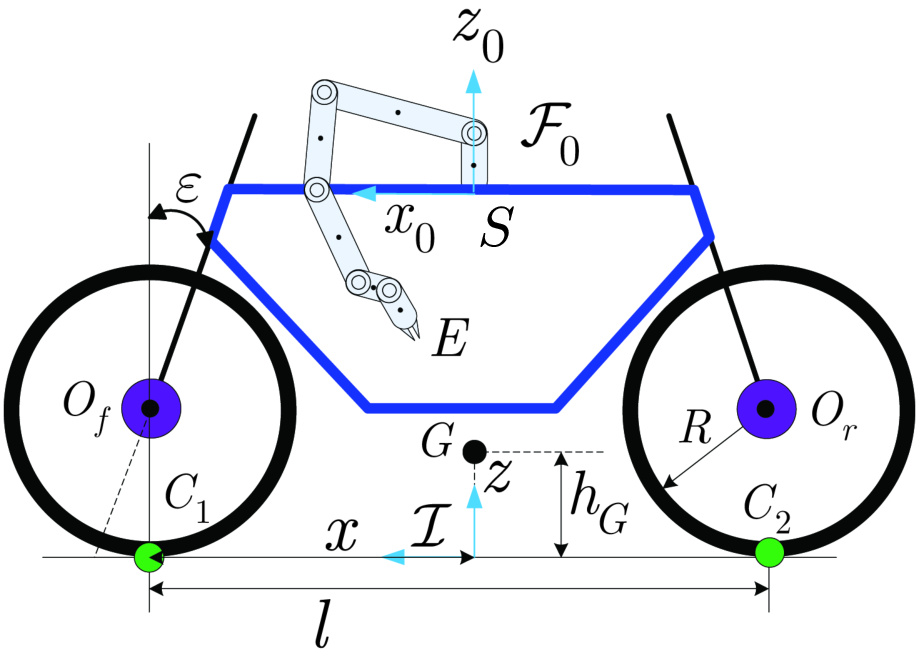}} 
	\hspace{1mm}
	\subfigure[]{
	\label{Fig_Bike_Arm_Schematics}
		\includegraphics[width=1.1in]{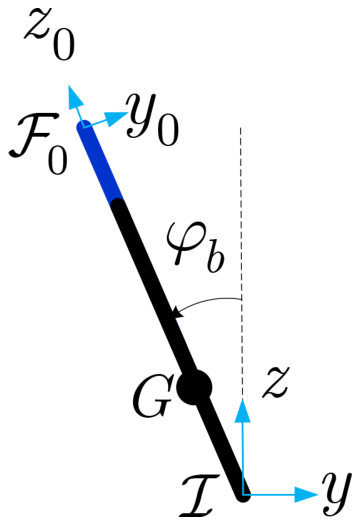}}
	\vspace{-0mm}
	\caption{(a) The prototype of the two-wheel steered bikebot mobile manipulation system. (b) Side-view configuration of the bikebot manipulation system. (c) Fron-view schematic of the bikebot roll motion.}
	\vspace{-0mm}
\end{figure*}

In this paper, we present a mobile manipulation system that is built on an autonomous bikebot. A 6-DOF lightweight manipulator is mounted on the bikebot and the system was developed for agricultural applications such as autonomous plant inspection and scouting~\cite{EdmondsCASE2019}. {All existing agricultural robots are built on double-track platform and their energy consumption is much higher than that of single-track mobile robots such as bikebot~\cite{Muetze2007bicycles}.} It is challenging for double-track robots to navigate in narrow, cluttered spaces and to actively probe and flexibly inspect objects under the canopy of densely-grown, tall plants. Light-weight bikebot provides additional advantages for small footprints that potentially avoids potential severe soil compaction~\cite{BecharBE2016}. Steering and speed control of autonomous one-wheel steered bikebot has been reported (e.g.,~\cite{WangICRA2017,Wang2020Stability}) but balance control of two-wheel steered bikebot for manipulation has not been studied. Because of the unstable platform and limited actuation, assistive devices were used to generate additional balance torque~\cite{Zhang2014Stationary,SEEKHAO2020102386Development,keo2011control,CuiIROS2020}. However, the additional balance actuators increase the systems complexity and operation cost. In this work, steering is used as the only actuation for balancing the stationary platform.

We focus on stationary balance of the bikebot manipulation for several reasons. First, it is more challenging to balance a stationary bikebot than a moving platform. With front wheel steering actuation, the bikebot can be only balanced within a small range of roll motion at stationary (i.e., 2-3 degs)~\cite{ZhangICRA2011,singhania2019study}. It is desirable to design new mechanisms and control methods to enlarge the controllable roll motion range for practical bikebot applications. Second, many applications such as plant inspection require that the mobile platform stays stationary, while the onboard manipulator  conducts the visual inspection or sample manipulation task. Therefore, robotic applications require the stationary balance capability of single-track mobile robots.

This paper presents the coordinated control of the bikebot and the manipulator to enhance the stationary balance and posture control task. The use of two-wheel steering and onboard manipulator enhances the balance capability. We first present a dynamic model of the system. A steering balance model is presented to analyze the steering configuration and maximize the balance capability. The balance condition is captured by an extended balance equilibrium manifold (BEM) of the mobile manipulation system. A BEM-enabled coordinated trajectory planning and control design is presented to achieve a balance-prioritized posture control. We conduct extensive experiments to validate and demonstrate the performance of the mechatronic and control design. The main contributions of this work are twofold. First, the presented two-wheel steering actuation analysis and model are innovative and provide a guidance on how to use the two-wheel steering design to increase balance capability of single-track mobile robots. It further explains the steering-induced balance capability differences between the single-track robot such as bicycles and other two-wheel, double-track Segway-like balance robots. Second, the proposed coordinated motion control design integrates the dynamic balance requirements with the task priority-based planning of a kinematic redundant manipulator. The extended BEM provides a new control approach to integrate the dynamic and kinematic constraints for mobile manipulation.

The remainder of this paper is organized as follows. Section~\ref{Section_Modeling} presents the problem statement and the systems dynamics. In Section~\ref{Section_Steering_Mechanism}, we analyze the two-wheel steering mechanism and discuss the balance torque model. Section~\ref{Section_Coordinated Planning} presents the coordinated pose control of the mobile manipulation. Experimental results are presented in Section~\ref{Section_Experiment} before we summarize the concluding remarks in Section~\ref{Section_Conclusion}.

\section{Problem Statement and Systems Dynamics}
\label{Section_Modeling}

In this section, we first present the problem statement and then the dynamic models of the bikebot manipulation system.

\vspace{-1mm}
\subsection{System Configuration and Problem Statement}

Fig.~\ref{Fig_Bike_Photo} shows the prototype of the bikebot manipulation. Fig.~\ref{Fig_Schemetics} illustrates the side-view schematic of the kinematic configuration and Fig.~\ref{Fig_Bike_Arm_Schematics} for a front-view of the system. An $n$-link lightweight manipulator with end-effector $E$ is mounted on the bikebot body frame at point $S$. The two wheel/ground contact points are denoted as $C_1$ and $C_2$ and the wheelbase is $l$. Two sets of coordinate frames are introduced: an inertial frame $\mathcal{I}$ and body frame $\mathcal{F}_i$ for the $i$th manipulator link, $i=1,\cdots,n$. Frame $\mathcal{F}_i$ is constructed by following the DH parameter convention~\cite{corke2017robotics}. $\mathcal{F}_0$ and $\mathcal{F}_n$ are for the base (platform) and end-effector frames, respectively. The horizontal and vertical distances from the bikebot's mass center $G$ to $C_1$ are $l/2$ and $h_G$, respectively. The front and rear steering mechanisms are symmetric with same caster angle $\varepsilon$.

The bikebot's steering and roll angles are denoted as $\phi$ and $\varphi_b$, respectively. For both front and rear steering angles, the positive direction is defined as the counterclockwise about the steering axis. We define $\bs \Theta=[\theta_1 \,\cdots\, \theta_n]^T$ as the manipulator joint angles, $i=1,\cdots,n$. The generalized coordinates of the system are denoted as $\bs q=[\varphi_b \; \bs \Theta^T]^T \in \mathcal{Q} \subset \mathbb{R}^{n+1}$, where $\mathcal{Q}$ is admissible set for $\bs{q}$. We denote the pose (i.e., position and orientation) of end-effector $E$ in $\mathcal{I}$ as $\bs\xi_e \in \mathbb{R}^6$.

{\em Problem Statement}: Given a set of ${N_\xi}$ desired poses $\{\bs\xi_{e}^k\}_{k=1}^{N_\xi}$, ${N_\xi} \in \mathbb{N}$, the goal is to design a planning and control method for the bikebot manipulation (i.e., steering and joint angles control) to let end-effector $E$ go through and hold stationary for short time at each $\bs\xi_{e}^k$, $k=1,\cdots,N_\xi$.

\subsection{Systems Dynamics}

We use DH parameters $(\theta_i, d_i, a_i, \alpha_i)$ for the $i$th link of the manipulator, $i=1,\cdots, n$. The homogeneous transformation matrix from $\mathcal{F}_i$ to $\mathcal{F}_0$ is written as~\cite{Murray1994}
\begin{equation}
	\label{Eq_Arm_Forward_Kine}
	\mathcal{T}_i(\bs q)= \mathcal{A}^0_1 \mathcal{A}^1_2\cdots \mathcal{A}^{i-1}_i,
\end{equation}
where $\mathcal{A}^{i-1}_i$ denotes the transformation from $\mathcal{F}_{i}$ to $\mathcal{F}_{i-1}$ as
\begin{equation}
	\label{Eq_Homegeneous_Transformation}
	{{\cal A}^{i-1}_i} = \begin{bmatrix} \bs{R}^{i-1}_i& \bs{p}_i \\  \bs{0}_{1\times 3} & 1
	\end{bmatrix},
\end{equation}
$\bs{R}^{i-1}_i=\mathbf{R}_z(\theta_i)\mathbf{R}_x(\alpha_i)$, $\mathbf{R}_j(\beta) \in \rm SO(3)$, $j=x,y,z$, denotes the rotational matrix about $j$-axis with angle $\beta$, $\bs p_i= [a_i\cos{\theta_i}\;a_i\sin{\theta_i}\;d_i]^T$ is the corresponding position vector in $\mathcal{F}_{i-1}$. With~(\ref{Eq_Arm_Forward_Kine}), we write the pose of end-effector $E$ in $\mathcal{F}_{0}$ as $\bs\xi^{\mathcal{F}_{0}}_e=\bs{\xi}_e(\mathcal{T}_n(\bs q))$.

We denote the mass center position of the $i$th link in $\mathcal{F}_0$ as $\bs{p}_{i_c}^{0}$ and its position in $\mathcal{I}$ is $\bs p_{i_c}=\bs R_0^{\mathcal{I}}(\bs{p}_0+\bs{p}^{0}_{i_c})$, where $\bs p_0$ is the position vector of point $S$ in $\mathcal{I}$. The linear velocity $\bs{v}^{0}_{i_c}$ and angular velocity $\bs{\omega}^{0}_{i_c}$ of the $i$th link in $\mathcal{F}_0$ are obtained as
\begin{equation}
\label{Eq_LinkCenterSpeed_InF0}
    \dot{\bs{\xi}}_{i_c}^{\mathcal{F}_{0}}=\left[\left(\bs{v}^{0}_{i_c}\right)^T \; \left(\bs{\omega}^{0}_{i_c}\right)^T\right]^T=\bs{J}_{i_c}\dot{\bs{\Theta}},
\end{equation}
where $\bs{J}_{i_c} \in \mathbb{R}^{6\times n}$ is the Jacobian from $\mathcal{F}_{i}$ and $\mathcal{F}_{0}$. Therefore, the linear velocity $\bs v_{i_c}$ and angular velocity ${ \bs \omega}_{i_c}$ in $\mathcal{I}$ are
\begin{equation}
\label{Eq_LinkCenterSpeed_Inertia}
    \bs v_{i_c}=\bs \omega_b\times \bs R_{0}^{\mathcal{I}}({ \bs p_0+ \bs p^{0}_{i_c}}) + \bs R_{0}^{\mathcal{I}} \bs v^{0}_{i_c}, {\bs \omega}_{i_c}= \bs \omega_b+\bs R_{0}^{\mathcal{I}} \bs \omega_{i_c}^0,
\end{equation}
where $\bs \omega_b=[\dot{\varphi}_b \; 0 \; 0]^T$ is the platform's roll velocity.

The dynamic model of the mobile manipulation system is obtained through Lagrange's equations. The system's kinetic and potential energies are
\begin{equation}
\label{Eq_Kinetic_Energy}
T=T_b+\sum\nolimits_{i=1}^{n} T_i, \; U =U_b+\sum\nolimits_{i=1}^{n}{U}_i,
\end{equation}
where $T_b= \frac{1}{2}\bs{\omega}_b^T \bs{I}_b \bs{\omega}_b+\frac{1}{2}m_b  \bs{v}_G^T \bs{v}_G$ is the kinetic energy for the bikebot and for the $i$th link of the manipulator $T_i= \frac{1}{2}m_{i}\bs{v}_{i_c}^T \bs{v}_{i_c}+ \frac{1}{2}\bs{\omega}_{i_c}^T \bs{R}_i^{\mathcal{I}} \bs{I}_i (\bs{R}_i^{\mathcal{I}})^T \bs{\omega}_{i_c}$, $m_b$ and $m_i$ are respectively the masses for the bikebot and the $i$th link, ${\bs{v}}_G$ is the velocity of the mass center $G$, $\bs I_b=\diag(I_b, 0, 0)$ and $\bs{I}_i$ are the inertia matrices for the bikebot and the $i$th link about their mass center, respectively. For potential energy terms in~(\ref{Eq_Kinetic_Energy}), for the bikebot, $U_b= {m_b}g\left( {\bs p_G \cdot {\bs e_z} + \Delta h_G} \right)$ and for the $i$th link, ${U}_i = {m_i}g {\bs p_{i_c} \cdot {\bs e_z}}$, where $\bs p_G$ is the position vector of $G$ in $\mathcal{I}$, $g=9.8$ m/s$^2$ is the gravitational constant, unit vector $\bs e_z=[0 \; 0 \; 1]^T$, and $\Delta h_G$ is the height change of $G$ due to steering actuation~\cite{Gong2019Control}.

The dynamic model is obtained by Lagrangian method as
\begin{displaymath}
  \bs D\left(\bs q \right)\ddot {\bs q} + \bs C\left( {\bs q,\dot{\bs q}} \right)\dot{\bs q} + \bs G\left( \bs q \right) =\bs \tau,
\end{displaymath}
where $\bs D \left( \bs q \right)\in \mathbb{R}^{(n+1)\times(n+1)}$, $\bs C\left( {\bs q,\dot{\bs q}} \right) \in \mathbb{R}^{(n+1)\times(n+1)}$, and $\bs G\left( \bs q \right) \in \mathbb{R}^{n+1}$ are the inertia, Coriolis, gravitational matrices, respectively. We omit the details for these lengthy matrices. The generalized force $\bs \tau=[\tau_{b} \;  \bs{\tau}^T_{\theta}]^T\in \mathbb{R}^{n+1}$ includes the controlled steering-induced balance torque $\tau_{b}$ and joint torque vector $\bs{\tau}_{\theta} \in \mathbb{R}^n$ for the manipulator. We further write the above model into following block matrix form
\begin{equation}
\label{Eq_Dynamics_Bike_Arm}
\begin{bmatrix} D_{bb} & \bs D_{b\theta} \\ \bs D_{\theta b} & \bs D_{\theta \theta} \end{bmatrix}
\begin{bmatrix} \ddot \varphi_b \\ \ddot {\bs \Theta} \end{bmatrix}+\begin{bmatrix}  \bs{C}_{b} \\
  \bs C_{\theta} \end{bmatrix} \dot{\bs q} +\begin{bmatrix}  G_{b} \\  \bs G_{\theta} \end{bmatrix}
=\begin{bmatrix} \tau_{b} \\  \bs \tau_{\theta} \end{bmatrix},
\end{equation}
where the block matrices are in appropriate dimensions and their dependencies on $\bs q$ and $\dot {\bs q}$ are dropped for presentation brevity. We will derive the steering-induced torque model for $\tau_b$ in the next section.

\section{Steering Balance Model}
\label{Section_Steering_Mechanism}

\begin{figure}[h!]
\hspace{-4mm}
\subfigure[]{
\label{model1}
\includegraphics[width=2.5in]{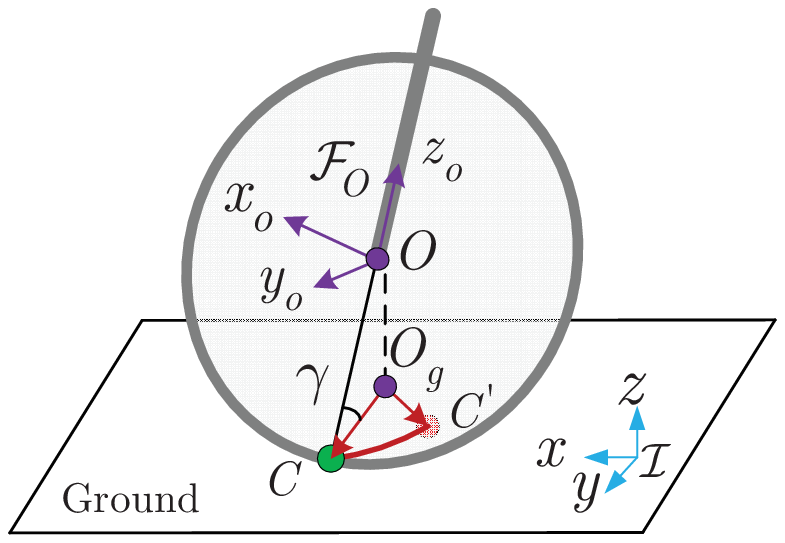}}
\hspace{-6mm}
\subfigure[]{
	\label{model2}
	\includegraphics[width=1.15in]{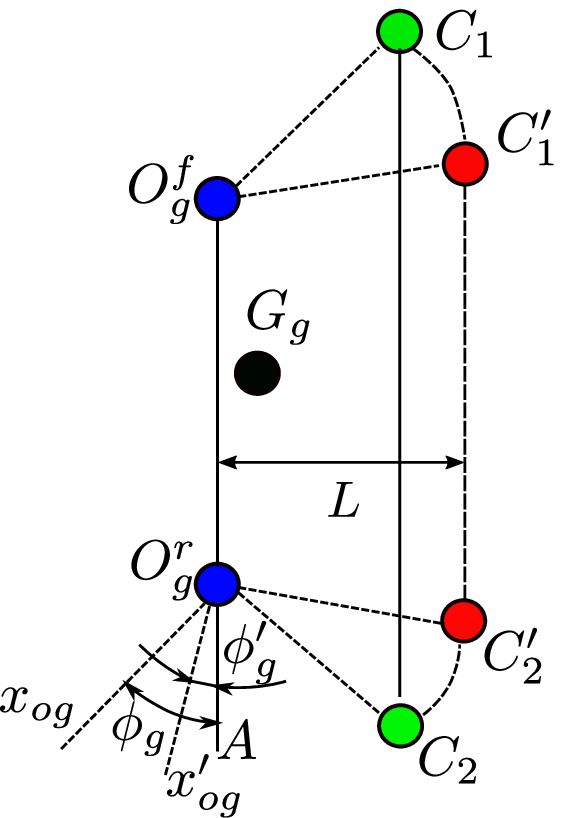}}
  \caption{Illustration of the steering mechanism and analysis. (a) Wheel plane geometry under steering angle increment $\delta$. Wheel contact point changes from $C$ to $C'$. (b) Wheel contact points $C_1C_2$ move to $C'_1C'_2$ under a small steering angle increment $\delta$ and the geometric relationships between $G_g$ and wheelbase line $C_1'C_2'$.}
  \label{Fig_Steering_Mechanism}
\vspace{-2mm}
\end{figure}

In this section, we analyze the steering mechanism and derive a model to obtain the configuration that produces maximum steering-induced balance torque. Fig.~\ref{model1} illustrates the schematic of the steering effect. We denote the wheel frame as $\mathcal{F}_O$ with wheel center $O$ and the $z_O$-axis is along the steering axis and the $y_O$-axis is perpendicular to the wheel plane. The projection of $O$ on the ground is denoted as $O_g$ and the wheel/ground contact point as $C$. We consider quasi-static steering motion such that steering angle $\phi$ is built on an initial steering angle $\phi_0$ with a small increment $\delta$, namely, $\phi=\phi_0+\delta$. Increment $\delta$ is small and the position change of point $G$ under $\delta$ is negligible.

The orientation of the wheel plane with respect to frame $\mathcal{I}$ is approximately obtained by three successive rotations: first $-\phi$ about $z_O$-axis, then $-\varphi_b$ about $x_O$, and finally $-\varepsilon$ about $y_O$-axis. With this observation, we obtain the rotational transformation from $\mathcal{F}_O$ to $\mathcal{I}$ as
\begin{equation}
\bs R_{\mathcal{F}_O}^{\mathcal{I}}=\mathbf{R}_y(-\varepsilon)\mathbf{R}_x(-\varphi_b)\mathbf{R}_z(-\phi).
\label{eq0}
\end{equation}
We denote the angle between the wheel plane and the ground as $\gamma$ and it is straightforward to obtain
\begin{equation}
\label{Eq_Steering_Radius1}
\cos  \gamma  = \bs R_{\mathcal{F}_O}^{\mathcal{I}} \bs e_y \cdot {\bs e_z}=\sin \phi \sin \varepsilon-\cos \phi \cos \varepsilon \sin {\varphi_b},
\end{equation}
where unit vector $\bs e_y=[0\;1\;0]^T$.

To simplify the two-wheel steering design, both the front and rear steering angles, denoted respectively by $\phi^f$ and $\phi^r$, are controlled and kept at symmetric position (i.e., same amplitude but opposite directions) for all time, namely, $\phi^f=-\phi^r=\phi$, with $\phi^f=\phi_{f0}+\delta$, $\phi^r=\phi_{r0}-\delta$ and $\phi_{f0}=-\phi_{r0}$. For brevity, we only use $\phi$ and $\delta$ in the following derivation. Under small $\delta$, wheel/ground contact points $C_1$ and $C_2$ move to $C'_1$ and $C'_2$, respectively. Fig.~\ref{model2} illustrates the wheel/ground contact points under $\delta$. With the above configuration, points $C_1$ and $C'_1$ ($C_2$ and $C'_2$) are located on a circular arc that is centered around $O^f_g$ ($O^r_g$), projected points $O^f$ ($O^r$) on the ground. The radii of the circular arc $\wideparen{C_1C'_1}$ ($\wideparen{C_2C'_2}$) and the bikebot wheel are denoted as $r$ and $R$, respectively. From the geometric relationship, we obtain
\begin{equation}
\label{Eq_Steering_Radius2}
r=R\cos\gamma= R\sin \phi \sin \varepsilon  - R\cos \phi \cos \varepsilon \sin {\varphi _b}.
\end{equation}

As shown in Fig.~\ref{model2}, under the same front and rear steering angles, wheelbase lines $C_1C_2$ and $C_1^\prime C_2^\prime$ are parallel. The corresponding projected steering angles are $\phi_g=\angle AO_g^r x_{og}$ and $\phi'_g=\angle A O_g^r x'_{og}$ (for the rear wheel). Let $L$ denote the distance from the $C_1 C_2$ to $O_g^fO_g^r$ and $L$ is obtained by the geometry relationship as
\begin{equation}
	\label{Eq_L}
L=r \cos \phi_{g}.
\end{equation}
The relationship between $\phi_{g}$ and $\phi$ is captured by \cite{Gong2019Control}
\begin{equation}\label{Eq_Steering_Projection}
  \phi_{g}=\arctan\left(\frac{\cos \varepsilon}{\cos {\varphi_b}}\tan\phi\right).
\end{equation}
Given a fixed $\phi_0$, from~(\ref{Eq_Steering_Radius2}), $r=r_{\phi_0}$ is considered a constant value for a small roll angle (e.g., $\varphi_b\approx 0$) and therefore, plugging~(\ref{Eq_Steering_Radius2}) and~(\ref{Eq_Steering_Projection}) into~(\ref{Eq_L}), we obtain
\begin{equation*}
    L=r_{\phi_0} \frac{\cos \varphi_b}{\sqrt{\cos^2 \varphi_b+\cos^2\varepsilon \tan^2 \phi}}=L\left(\varphi_{b}, \phi_{0}, \delta\right),
\end{equation*}
where $L$ is considered as a function of $\varphi_{b}$, $\phi_{0}$ and $\delta$.

We approximate the gravity-induced balance torque by using the above calculated $L$ as the distance between $G_g$ (the projected point of $G$ on the ground) to $C'_1C'_2$ and obtain
\begin{equation}
	\label{EqTorq}
\tau_{b} \approx m_{b} g L=mg L\left(\varphi_{b}, \phi_{0}, \delta\right).
\end{equation}
It is helpful to find steering angle $\phi_0$ at which the increment $\delta$ generates largest torque increase of $\tau_b$. Thus, we introduce the steering torque sensitivity as the steering torque rate with respect to $\delta$ at $\varphi_{b}=0$, namely,
\begin{equation*}
\label{Eq_Steering_Sensitivity}
S_\tau(\phi_0) = \left|\frac{\partial \tau_b}{\partial \delta}\right|_{\substack{\delta=0 \\ \varphi_b= 0}} =m_bg r_{\phi_0}\frac{\cos^2\varepsilon\tan\phi_0(\tan^2 \phi_0 + 1)}{\left(\cos^2\varepsilon\tan^2 \phi_0 + 1\right)^{3/2}}.
\end{equation*}
From above equation, it is straightforward to see that at $\phi_0=0$, $S_\tau(\phi_0)=0$ and this implies that the commonly used zero steering angle has the minimum steering torque sensitivity.

We further calculate that at $\phi^*_0=\frac{\pi}{2}$ rad, $S_\tau(\phi_0)$ reaches its maximum value as
\begin{equation*}
S_\tau(\phi_0^*)= m_bg r_{\phi_0^*}\frac{1}{\sqrt{\cos^2 \varepsilon}} = m_bgR\tan \varepsilon.
\end{equation*}
Therefore, we focus on using $\phi_0^*=\frac{\pi}{2}$ rad for mobile manipulation control since it generates the largest balance torque per unit of steering angle. In this case, $\phi=\frac{\pi}{2}+\delta$ and $\phi_g=\frac{\pi}{2}+\delta_g$, the steering torque is then calculated as
\begin{equation}
\label{Eq_90_Steering_Torque}
\text{\hspace{-3.6mm}} \tau_{b90} = m_bg r_{90} \cos \phi_g= - m_bgR\sin \varepsilon\cos \delta \sin \left(\tfrac{\delta}{\cos \varepsilon}  \right),
\end{equation}
where $r_{90}=R \sin \varepsilon \cos \delta $ is from~(\ref{Eq_Steering_Radius2}) with $\varphi_b \approx 0$ and $\delta_g  \cos \varepsilon \approx \delta$ is taken from~(\ref{Eq_Steering_Projection}). It is clear that a large caster angle configuration helps increase the steering-induced balance torque and therefore improve balance capability. For any other initial steering angle $\phi_0$, the radius is calculated by~(\ref{Eq_Steering_Radius2}) and the steering torque $\tau_b$ is obtained by~(\ref{EqTorq}).

It is interesting to note that under $\phi_0=\frac{\pi}{2}$, the steering configuration is different with commonly used zero initial steering angle $\phi_0=0$. Indeed, the configuration is similar to double-track balance robot such as Segway. This observation explains and implies that double-track steering configuration such as Segway-like robots helps provide more steering-induced balance torques than the single-track configuration such as bicycles. We therefore use $\phi_0=\frac{\pi}{2}$ in the bikebot balance implementation.

\section{Coordinated Balance Control Design}
\label{Section_Coordinated Planning}

In this section, we focus on the coordinated balance and pose control of the bikebot manipulation. The controller design is built on the balancing equilibrium manifold (BEM).

\subsection{Balance Equilibrium Manifold}

The kinematics redundancy of the multi-DOFs manipulator enables the end-effector to reach the target poses with the balanced  bikebot platform. If we consider the manipulator moves quasi-statically (i.e., slowly), the balanced bikebot roll angle $\varphi_b$ and manipulator joint angles $\bs{\Theta}$ should satisfy an intrinsic relationship that is captured by BEM. From~(\ref{Eq_Dynamics_Bike_Arm}), the equation of motion of the bikebot is written as
\begin{equation}
\label{Eq_Bike_Roll}
D_{bb}\ddot{\varphi}_b+\bs D_{b\theta}\ddot{\bs \Theta}+\bs{C}_{b}\dot{\bs{q}}+G_{b}(\bs{q})=\tau_{b}
\end{equation}
where $G_b(\bs{q})$ is the total gravitational torque from the bikebot and the manipulator. Considering the quasi-static motion, namely, $\ddot{\bs{q}}=\dot{\bs{q}}=0$, we define the BEM as
\begin{equation}
\label{Eq_Equilibrium_Manifold}
    \mathcal{E}=\{\bs q_e=[\varphi^e_b \; \bs \Theta^T_e]^T: G_{b}(\bs{q})=\tau_b,\, \bs q\in \mathcal{Q}\}.
\end{equation}
The BEM captures all configurations that satisfy the static equilibrium constraint. Using BEM, we estimate the static maximum roll angle $\varphi^{\max}_b$ under the maximum balance steering $\tau^{\max}_b$ with possible $\bs{\Theta}$.

To move the end-effector from one pose to another, a trajectory should be designed around the BEM at any time, namely, $\bs q \in \mathcal{E}$. A velocity constraint should be enforced given the BEM and limited steering actuation. Using~(\ref{Eq_90_Steering_Torque}), the steering torque is $\tau_b= -MgR \sin  \varepsilon \cos  \delta \sin \delta_g$, where $M=m_b+\sum_{i=1}^n m_i$ is the total mass of the entire system. Taking derivative of BEM condition $G_b(\bs q)=\tau_b$, we obtain
\begin{equation*}
\label{Eq_D_SBC}
\dot{G}_b = \frac{{\partial {G_b}}}{{\partial \bs q}}\dot{\bs q} = -Mg R \sin  \varepsilon \frac{d}{d \delta }(\cos  \delta\sin \delta_g) \dot\delta=h(\delta) \dot\delta,
\end{equation*}
where $h(\delta)=-Mg R \sin  \varepsilon \frac{d}{d \delta }(\cos  \delta \sin \delta_g)$. Defining $\bs J_G=\frac{{\partial {G_b}}}{{\partial \bs q}}$ as a Jacobian-like matrix, the above velocity constraint is specified as
\begin{equation}
	\label{Eq_Speed_Constraint}
  |\bs J_{G}\dot {\bs q}| \leq h_{\max}\dot \delta_{\max},
\end{equation}
where $h_{\max}=\sup_{\delta} |h(\delta)|$ and $\dot{\delta}_{\max}$ is the maximum steering angular rate. Constraint~(\ref{Eq_Speed_Constraint}) implies that when designing the trajectory $\bs q(t)$, the allowed motion velocity is restricted by the steering angular rate.

\subsection{Balance-Prioritized Pose Trajectory Planning}
The end-effector pose workspace in $\mathcal{I}$ is defined as
\begin{equation}
\label{Eq_Manipulator_Workspace}
\mathcal{X}(\bs q)=\{\bs \xi_e: \, \bs \xi_e = \bs \xi_e(\mathcal{T}_{n+1}(\bs q)), \bs q \in \mathcal{E}, |\tau_b|\leq {\tau}_b^{\max}\},
\end{equation}
where $\mathcal{T}_{n+1}(\bs q)$ is the homogeneous transformation from $\mathcal{F}_n$ to $\mathcal{I}$. We further define the local end-effector pose workspace $\mathcal{X}_{\varphi_b^0}(\bs \Theta) \subseteq \mathcal{X} (\bs q)$ under roll angle $\varphi_b^0$, $\bs q^0=[\varphi_b^0 \, \bs{\Theta}^T]^T$,
\begin{equation*}
\label{Eq_Local_Wrokspace}
\mathcal{X}_{\varphi_b^0}(\bs \Theta)= \left\{ \bs{\xi}_e: \bs \xi_e= \bs \xi_e(\mathcal{T}_{n+1}(\bs q^0)), \bs q^0 \in \mathcal{E} \right\}.
\end{equation*}
The rationale to introduce $\mathcal{X}_{\varphi_b^0}(\bs \Theta)$ is to specify the bikebot roll angle $\varphi_b^0$ first for balance task and then use the manipulator to achieve the pose control task. We consider the task priority from high to low as follows: (1) bikebot platform balance; (2) pose control of end-effector $E$; and (3) collision avoidance during arm movement from one desired pose to another. Due to the redundant kinematics of the manipulator, we use a task-priority optimization approach to plan the trajectory.

We define the following balance-prioritized inverse kinematics (BPIK) problem. Given a sequence of desired end-effector poses $\bs \xi_{e}^k$, $k=1,\cdots,{N_\xi}$, the BPIK is to find optimal bikebot roll and manipulator joint angles $\bs q_k^*$ as
\begin{subequations}
\begin{align}
\bs q^*_{k}  =& \arg \mathop {\min }\limits_{\bs q} {\lambda _1}\Gamma_1 +\lambda _2\Gamma_2+ {\lambda _3}\Gamma_3 \label{Eq_Inverse_Kine:a}\\
\text{Subj. to} & \;\, \lambda_4 |G_b| \leq {\tau}_b^{\max},\; \bs q\in \mathcal{E}, \label{Eq_Inverse_Kine:b}
\end{align}
\label{Eq_Inverse_Kine}
\end{subequations}
\hspace{-1.6mm}where $\Gamma_1=\|\bs \xi_e^k-\bs \xi_e(\mathcal{T}_{n+1}(\bs{q}))\|_2^2$, $\Gamma_2=|G_b(\bs{q}_k)-G_b(\bs{q}_{k-1}^*)|^2$, $\Gamma_3= \bs e_{k-1}^T\bs P \bs e_{k-1}$, $\bs e_{k-1}=\bs q - \bs q_{k-1}^*$, $\lambda_1, \lambda_2, \lambda_3 >0$, and $\lambda_4>1$ are weight parameters, $\bs P>0$ is a symmetric positive definite matrix. We initialize with $\bs q_{-1}^*=\bs 0$ and $G_b(\bs{q}_{-1}^*)=0$.

Inequality~(\ref{Eq_Inverse_Kine:b}) is set as a hard constraint such that the steering output is always within the balance capability, while the pose regulation becomes a part of the objective function (i.e., $\Gamma_1$). Therefore, balance serves as a higher priority than pose regulation by the BPIK design. This is similar to the approach in~\cite{SIMETTI2019103287} by projecting the low level priority task into subspace of solution of high level tasks. Terms $\Gamma_2$ and $\Gamma_3$ in~(\ref{Eq_Inverse_Kine:a}) try to minimize the difference between configurations at the current and the previous steps. We use the BPIK to obtain $\bs q^*_k$ from $\bs \xi_e^k$. We first search the solution in local workspace $\mathcal{X}_{\varphi_b^0}(\bs \Theta)$ to avoid large bikebot movement. If this is impossible, the BPIK then searches the solution in the workspace $\mathcal{X}(\bs q)$. If the calculated feasible poses are outside of $\mathcal X(\bs q)$,~(\ref{Eq_Inverse_Kine}) will return the closest results. Once obtaining the desired joint angles $\{\bs q_k^*\}_{k=1}^{N_\xi}$, we need to design transition trajectory along $\mathcal E$ between each two consecutive poses.

We consider a pair of desired consecutive configurations $(\bs q^*_{k-1},\bs q^*_{k})$ to position end-effector $E$. With user-specified starting and ending times denoted respectively as $t_0$ and $t_f$, we define $\bs q(t_0) = \bs q^*_{k-1}$ and $\bs q(t_f) = \bs q^*_k$. A motion trajectory $\bs q^*(t)$ needs to be designed from $\bs q(t_0)$ and $\bs q(t_f)$ along $\mathcal E$. The trajectory planning is formulated as the following optimization problem.
\begin{subequations}
	\begin{align}
    \text{\hspace{-1mm}} \min_{\bs{q}(t)} \int_{{t_0}}^{{t_f}} & \bs e_{k-1}^T\bs W_1 \bs e_{k-1} +{\dot{\bs q}^T}\bs W_2\dot{\bs q} + (\delta G_{b,k})^2 dt\\
    \text{\hspace{-4mm} Subj. to} & \;\, \dot{\bs q}(t_0)=\dot{\bs q}^*_{k-1}, \dot {\bs q}(t_f) = \dot{\bs q}^*_{k}, \nonumber \\
    &  \ddot{\bs q}(t_0)=\ddot{\bs q}^*_{k-1},\ddot{\bs q}(t_f) =\ddot{\bs q}^*_k, \label{Eq_Trajectory_Plan:a}\\
    &\bs D_{\theta b}\ddot{\varphi}_b+\bs D_{\theta\theta}\ddot{\bs \Theta}+\bs C_{\theta}\dot{\bs{q}}+\bs G_{\theta}=\bs \tau_{\theta},\label{Eq_Trajectory_Plan:b}\\
    &|\bs J_{G}\dot {\bs q}|\leq h_{\max}\dot{\delta}_{\max}, \lambda_4 |G_b|\leq  {\tau}_b^{\max}, \label{Eq_Trajectory_Plan:c} \\
    &|\tau_{\theta,i}| \leq   {\tau}_{\theta,i}^{\max}, \bs q \in \mathcal{Q}, |\dot{\bs q}| \leq \dot{q}_{\max}, |\ddot{\bs q}| \leq \ddot{q}_{\max}, \label{Eq_Trajectory_Plan:d}
  \end{align}
\label{Eq_Trajectory_Plan}
\end{subequations}
\hspace{-3.5mm} where $\delta G_{b,k}=G_b(\bs q)-G_b(\bs q^*_{k-1})$, $\bs W_1, \bs W_2 \in \mathbb{R}^{n+1}$ are positive diagonal matrices, and $\tau_{\theta,i}^{\max}$ is the maximum joint torque of the $i$th link, $i=1,\cdots,n$. To consider the quasi-static motion, the angular velocity and acceleration of the manipulator are bounded as in~(\ref{Eq_Trajectory_Plan:d}). The constraint in~(\ref{Eq_Trajectory_Plan:c}) is similar to that in~(\ref{Eq_Inverse_Kine}) along with joint torque limits.

To solve~(\ref{Eq_Trajectory_Plan}), we try to avoid integration of the differential constraint~(\ref{Eq_Trajectory_Plan:b}) and B\'{e}zier polynomials are used to specify the solution in each dimension of $\bs q(t)$. We use B\'{e}zier polynomial because of its attractive properties~\cite{Wester2007}. The solution $\bs{q}(t)$ is written in term of $N$th order B\'{e}zier polynomials ($N\in \mathbb{N}$) as
\begin{equation*}
\label{Eq_Configuraoitn_Specification}
\varphi_b=b(s,\bs p_b)=\sum\limits_{j = 0}^N p_{b_j}b_j(s), \, \theta_i=b(s,\bs p_{\theta_i})=\sum\limits_{j = 0}^N p_{\theta_{ij}}b_j(s)
\end{equation*}
for $i=1,\cdots,n$, where $b_j(s)=\frac{N!}{(N - j)!j!}(1-s)^{N-j}s^j$, parameters $\bs p_b=[p_{b_0} \, \cdots\,  p_{b_N}]^T$ and $\bs p_{\theta_i}=[p_{\theta_{i0}} \, \cdots\,  p_{\theta_{iN}}]^T$. The normalized progress variable $s = \frac{t-t_0}{t_f-t_0}$ maps  $t\in[t_0, t_f]$ to $s \in [0,1]$.

From above formulation, we obtain $\bs{q}(t_0)=[b(0,\bs p_b) \, b(0,\bs p_{\theta_0}) \, \cdots \, b(0,\bs p_{\theta_n})]^T$, $\bs{q}(t_f)=[b(1,\bs p_b) \, b(1,\bs p_{\theta_0})\, \cdots \, b(1,\bs p_{\theta_n})]^T$ and $\bs{q}(t)$ is then written as polynomials of $s$ with parameters $\bs p=\{\bs p_b, \bs p_{\theta_1},\cdots, \bs p_{\theta_n}\}$. For $\dot{\bs{q}}(t)$, we obtain $\dot{\varphi}_b=\sum\limits_{j = 0}^N p_{b_j}b'_j(s)\frac{ds}{dt}$ and $\dot{\theta}_i=\sum\limits_{j = 0}^N p_{\theta_{ij}}b'_j(s)\frac{ds}{dt}$, $i=1,\cdots,n$. Noting that $\frac{ds}{dt}=\frac{1}{t_f-t_0}$ is constant, $\dot{\bs{q}}(t)$ and $\ddot{\bs{q}}(t)$ are written in term of $\bs p$. Therefore, the trajectory planning problem~(\ref{Eq_Trajectory_Plan}) is transformed into the $s$-domain and the differential constraints are written as algebraic formulation in polynomials of $s$ and $\bs p$.

To solve the optimization problem, we discretize $s \in [0,1]$ with $N_s$ sampling points and both the objective and constraint functions in~(\ref{Eq_Trajectory_Plan}) are evaluated at these points. A sequential quadratic programming (SQP) algorithm is then used to obtain the optimized trajectory $\bs{q}^*(t)$~\cite{nocedal2006sequential}. In each iteration, a total of $3N_s(n+1)$ evaluations of $\bs b(s; \bs p)$ (for $\bs q, \dot{\bs q}, \ddot{\bs q}$) and $N_s(n+ 2)$ evaluations of~(\ref{Eq_Trajectory_Plan:b}) and~(\ref{Eq_Trajectory_Plan:c}) are needed for $(n+1)(N-1)$ optimization variables. Additionally, the SQP solver has complexity $O(N^2n^2)$. Therefore, the computational complexity for solving~(\ref{Eq_Trajectory_Plan}) by the proposed B\'{e}zier polynomial-based approach is $O((N + N_s)Nn^2)$. As a comparison, a dynamic programming (DP) method can be used to solve~(\ref{Eq_Trajectory_Plan}) with complexity  $O(N^2_sn^2)$. Because of $N_s\gg N$, the proposed method is much faster than the DP method.

Algorithm~\ref{Algorithm_Trajectory_Generation} summarizes the trajectory planning as described above.

\begin{algorithm}[h!]
	\SetKwData{Left}{left}
	\SetKwData{This}{This}
	\SetKwData{Up}{up}
	\SetKwFunction{Union}{Union}
	\SetKwFunction{FindCompress}{FindCompress}
	\SetKwInOut{Input}{Input}
	\SetKwInOut{Output}{Output}
	Specify $\{\bs \xi_{e}^k\}_{k=1}^{N_\xi}$, $\lambda_i, i=1,...,4$, $\bs P$, $\bs W_1$, $\bs W_2$, and $\epsilon >0$\;
	\For{$k \le {N_\xi}$}{
		\If{$k=1$}{
		$\bs q^*_{k-1} \leftarrow \bs 0$, $G(\bs q^*_{k-1}) \leftarrow 0$\;
		Solve $\bs q^*_{k} \in \mathcal{X}(\bs q)$ by~(\ref{Eq_Inverse_Kine}) with $\bs \xi_{e}^k$\;
		}
	\Else{
			Solve $\bs q^*_{k} \in \mathcal{X}_{\varphi_{b,k-1}^*} (\bs \Theta)$ by~(\ref{Eq_Inverse_Kine}) with $\bs \xi_{e}^k$\;
			Calculate the pose $\bs \xi_e (\bs q^*_{k})$\;
			\If{$\|\bs \xi_{e}^k-\bs \xi_e (\mathcal{T}_{n+1}(\bs q^*_{k})) \| \ge \epsilon$}{
				Re-calculate $\bs q^*_{k} \in \mathcal{X}(\bs q)$ by~(\ref{Eq_Inverse_Kine})\;
			}
		}
		$k \leftarrow k+1$
	}
	\For{$ j \le {N_\xi}$}
	{Specify $t_0$ and $t_f$; $\bs q\left( {{t_0}} \right) \leftarrow \bs q^*_{j-1}$, $\bs q\left( {{t_f}} \right) \leftarrow \bs q^*_{j}$\;
Plan $\bs{q}^*(t)$, $t\in [t_0,t_f]$ by~(\ref{Eq_Trajectory_Plan}) using B\'{e}zier polynomial specification\;
$j \leftarrow j+1$ \;
	}
	\caption{Trajectory planning for pose regulation}
	\label{Algorithm_Trajectory_Generation}
\end{algorithm}

\subsection{Bikebot Steering and Manipulator Control}

\begin{figure}[htb!]
	\centering
	\vspace{1mm}
	\includegraphics[width=3.45in]{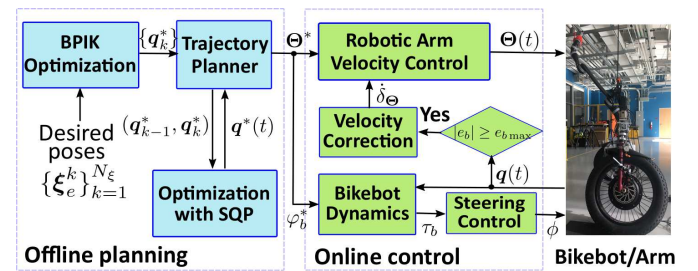}
	\caption{The block diagram of the balance-prioritized trajectory planning and control scheme.}
	\label{Fig_Control_Loop}
	\vspace{-1mm}
\end{figure}

Fig.~\ref{Fig_Control_Loop} illustrates the balance-prioritized trajectory planning and control design. The previous section discusses the trajectory planner to obtain $\bs{q}^*(t)$ for a given set of end-effector poses $\{\bs{\xi}_e^k\}_{k=1}^{N_\xi}$. In this subsection, we present the online control algorithms to follow the trajectory $\bs{q}^*(t)$.

We first present the steering control of the bikebot to follow $\varphi_b^*$ and then discuss the manipulator controller to follow $\bs{\Theta}^*$. Using the dynamic model of the bikebot in~({\ref{Eq_Bike_Roll}}), we define the roll angle error $e_b = \varphi_b-\varphi_b^*$ and a feedback linearization control is designed as
\begin{equation}
\label{Eq_Balance_Control_PID}
\tau_b = D_{bb}\ddot{\varphi}^*_b+\bs D_{b\theta}\ddot{\bs \Theta}+\bs{C}_{b}\dot{\bs{q}}+G_{b}+ k_p e_b+k_d \dot{e}_b,
\end{equation}
where $k_p, k_d>0$ are the feedback gains. From~({\ref{Eq_Balance_Control_PID}}), we use the steering balance model ({\ref{Eq_90_Steering_Torque}}) to obtain the steering angle $\phi$.

For the manipulator, since the trajectory are designed in a quasi-static form, we take the velocity control to follow the desired trajectory $\bs{\Theta}^*$ in $\mathcal{F}_0$. We recognize $\ddot {\bs\Theta}\approx 0$ after the system compensating the gravity ($G_\theta$). We extend the velocity control in~\mbox{\cite{RAJA2019103245,Antonelli2009Stability}} with additional velocity correction under bikebot roll motion errors. Defining joint angle error $\bs{e}_\Theta=\bs{\Theta}-\bs{\Theta}^*\in \mathbb{R}^n$, the velocity control in the joint workspace is given by
\begin{equation}
\dot{\bs{e}}_\Theta=-\bs{K}_p \bs{e}_\Theta+ I_\Theta \dot{\bs{\delta}}_{\Theta},
\label{velcontrol}
\end{equation}
where $\bs{K}_p=\diag\{K_{p1},\cdots,K_{pn}\}$ with $K_{pi}>0$, $i=1,\cdots,n$ and $I_\Theta=1$ if $|e_{b}|>\varepsilon_b$; otherwise $I_\Theta=0$, with an error threshold $\varepsilon_b>0$. The velocity correction $\dot{\delta}_{\bs\Theta}$ in~({\ref{velcontrol}}) is designed as
\begin{displaymath}	
\dot{\bs{\delta}}_{\Theta}=  - \kappa \tfrac{\partial \delta G_b}{\partial \bs \Theta } = 2\kappa \left[G_b(\bs q^*) - G_b(\bs q) \right]{\left(\tfrac{\partial {G_b}(\bs q)}{\partial \bs \Theta}\right)^T},
\end{displaymath}
where $\delta G_b= (G_b(\bs q^*)-G_b(\bs q))^2$ denotes deviation of actual motion from the BEM and $\kappa>0$ is a scalar.

Under~(\ref{Eq_Balance_Control_PID}), the closed-loop bikebot roll dynamics is
\begin{equation}
\ddot e_b+k_d\dot e_b+k_pe_b=0
\label{errodyn0}
\end{equation}
and roll error $e_b(t)$ converges to zero exponentially. Without loss of generality, let $k_d^2<4k_p$ and then from~(\ref{errodyn0}) we obtain
\begin{equation}
|e_b(t)| \leq M_b e^{-\frac{k_d}{2}t},
\label{errodyn1}
\end{equation}
where $M_b>0$ is a finite constant that is related to $e_b(0)$ and $\dot{e}_b(0)$. Therefore, for any $t\geq t_b:=\frac{2}{k_d} \ln \left(\frac{M_b}{\varepsilon_b}\right)$, $|e_b(t)|\leq \varepsilon_b$. To show the tracking error convergence for $\bs{e}_\Theta$, we consider a Lyapunov function candidate $V(t)=\bs{e}^T_\Theta \bs{e}_\Theta=\|\bs{e}_\Theta\|^2 > 0$ for any non-zero error $\bs{e}_\Theta$. Let $l_\Theta$ denote the upper bound for input $I_\Theta \dot{\bs \delta}_{\Theta} $ during $t\in [0,t_b]$, namely, $l_\Theta:=\sup_{0\leq t\leq t_b}\|I_\Theta \dot{\bs \delta}_{\Theta}\|$. Moreover, we have
\begin{equation*}
\dot{V}(t)=-2\bs e^T_\Theta \bs{K}_p \bs{e}_\Theta+2I_\Theta \bs e^T_\Theta \dot{\bs \delta}_{\Theta} \leq -2\lambda_p \|\bs{e}_\Theta\|^2+2l_\Theta \|\bs e_\Theta\|,
\label{Eq_V_dot}
\end{equation*}
where $\lambda_p=\min_{1\leq i\leq n} K_{pi}$. We introduce $W(t)=\sqrt{V(t)}=\|\bs{e}_\Theta(t)\|$ and from above inequality, we obtain $\dot W \leq -{\lambda_{p}}W+{l_\Theta}$. Thus, we have $\frac{d}{dt}(W e^{{\lambda_p}t})=\dot We^{{\lambda_p}t}+W {\lambda_p}e^{{\lambda_p}t} \leq  {l_\Theta}e^{{\lambda_p}t}$ and integrating from $0$ to $t$, we obtain
\begin{equation*}
W(t) e^{{\lambda_p}t}-W(0) \leq \tfrac{l_\Theta}{\lambda_{p}}(e^{{\lambda_p}t}-1).
\end{equation*}
Noting $W(t)=\|\bs{e}_\Theta(t)\|$, the above inequality becomes
\begin{eqnarray}
\|\bs{e}_\Theta(t)\| &\leq & \|\bs{e}_\Theta(0)\|e^{-{\lambda_p}t}+\tfrac{l_\Theta}{\lambda_{p}}(1-e^{-{\lambda_p}t}) \nonumber \\
&\leq& \|\bs{e}_\Theta(0)\|e^{-{\lambda_p}t}+\tfrac{l_\Theta}{\lambda_{p}}
	\label{EqError0}
\end{eqnarray}
From the above analysis, $\dot V$ is negative outside of the compact set $\mathcal{S}=\{\bs e_{\Theta}: \|\bs e_{\Theta}(t)\| \leq \frac{l_{\Theta}}{\lambda_{p}}\}$ and $\bs e_{\Theta}(t)$ exponentially converges to $\mathcal{S}$. Note that for $t \geq t_b$, $I_\Theta = 0$, $\bs{e}_\Theta(t)$ converges to zeros exponentially due to stable dynamics~(\ref{velcontrol}).
	
With error convergence in~(\ref{errodyn1}) and~(\ref{EqError0}), we obtain the error bound for $\bs{e}_q=\bs{q}-\bs{q}^*=[e_b \; \bs{e}^T_\Theta]^T$ as
\begin{eqnarray}
\|\bs{e}_q(t)\|&=&\sqrt{e^2_b(t)+\|\bs{e}_\Theta(t)\|^2}\leq|e_b(t)|+\|\bs{e}_\Theta(t)\| \nonumber \\
&\leq & M_b e^{-\frac{k_d}{2}t}+\|\bs{e}_\Theta(0)\|e^{-{\lambda_p}t}+\tfrac{l_\Theta}{\lambda_{p}}.
\label{EqError1}
\end{eqnarray}
Considering pose error $\bs{e}_\xi=\bs{\xi}_e(\bs{q})-\bs{\xi}_e(\bs{q}^*)$ and $\bs{q}=\bs{q}^*+\bs{e}_q$, we have
\begin{equation}
\bs{e}_\xi=\bs{\xi}_e(\bs{q}^*)+\frac{\partial \bs{\xi}}{\partial \bs{q}}\Big\vert_{\bs{q}^*}\bs{e}_q+\bs{\Delta}_q-\bs{\xi}_e(\bs{q}^*)=\bs{J}_e(\bs{q}^*)\bs{e}_q+\bs{\Delta}_q,
	\label{EqError2}
\end{equation}
where $\bs{J}_e(\bs{q}) \in \mathbb{R}^{6\times(n+1)} $ is the Jacobian matrix from $\mathcal{F}_n$ (end-effector $E$) to inertial frame $\mathcal{I}$ and $\bs{\Delta}_q \in \mathbb{R}^6$ is the higher order term of error $\bs{e}_q$. From~(\ref{EqError2}), it is straightforward to obtain that $\|\bs{e}_\xi(t)\|\leq \|\bs{J}_e(\bs{q}^*)\|\|\bs{e}_q(t)\|+\|\bs{\Delta}_q\|$. For the higher order term $\bs{\Delta}_q=O(\|\bs{e}_q\|^2)$, there exists a finite constant $M_\delta>0$ such that $\|\bs{\Delta}_q\| \leq M_\delta \|\bs{e}_q(t)\|$ due to~(\ref{EqError1}) and then $\|\bs{e}_\xi(t)\|\leq M_q \|\bs{e}_q(t)\|$, where $M_p=\sup_{\bs{q}^*} \|\bs{J}_e(\bs{q}^*)\|+M_\delta$. Therefore, the pose error $\bs e_\xi(t)$ converges to a small ball near zero exponentially and the robotic system is stable.

\section{Experiments}
\label{Section_Experiment}

\subsection{Experiment Setup}

\setcounter{figure}{5}
\begin{figure*}[htb!]
	\centering
	\subfigure[]{
		\label{Fig_Steering_Sensitivity:a}
		\includegraphics[width=2.38in]{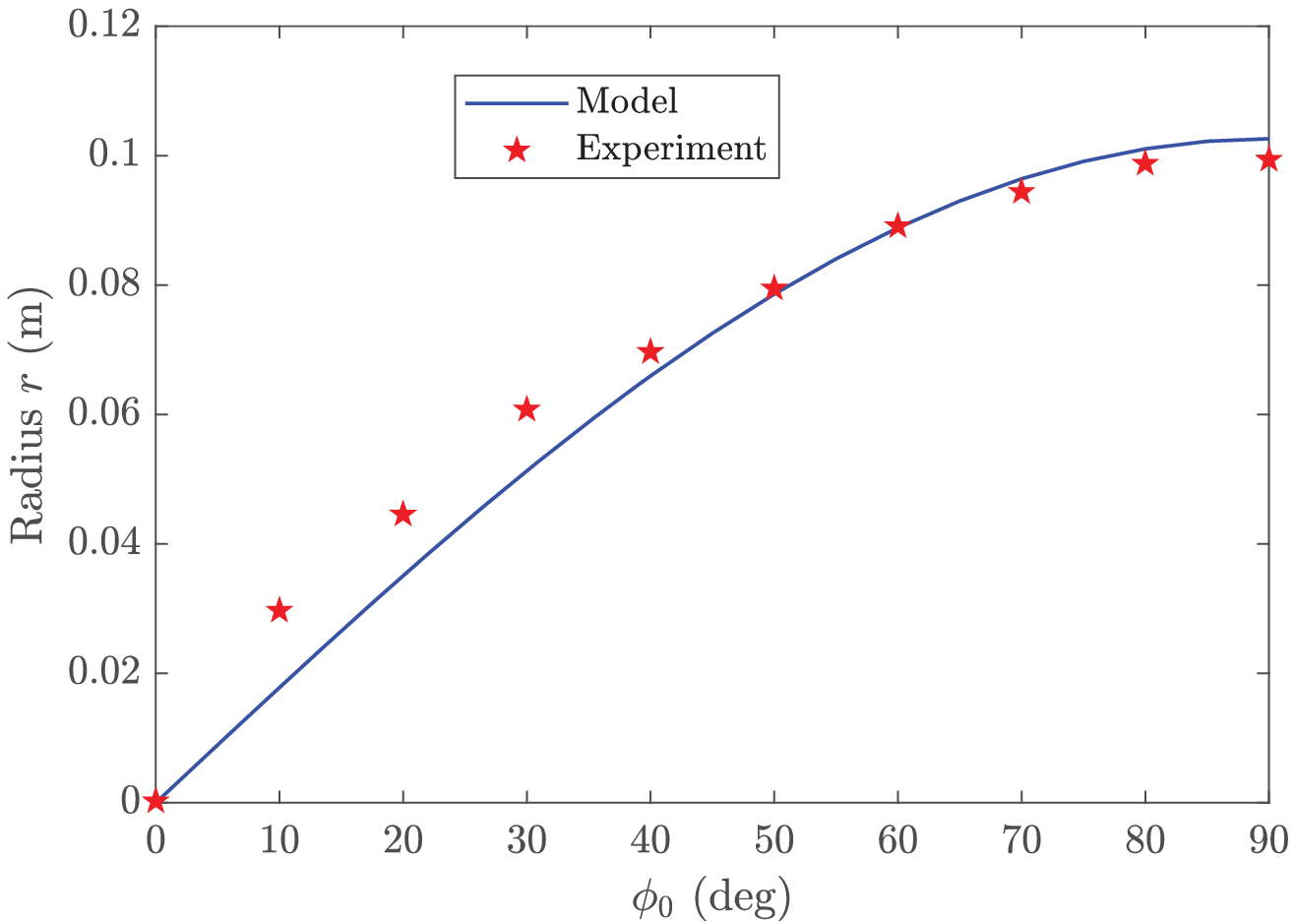}}
	\hspace{-4.5mm}
	\subfigure[]{
		\label{Fig_Steering_Sensitivity:b}
		\includegraphics[width=2.36in]{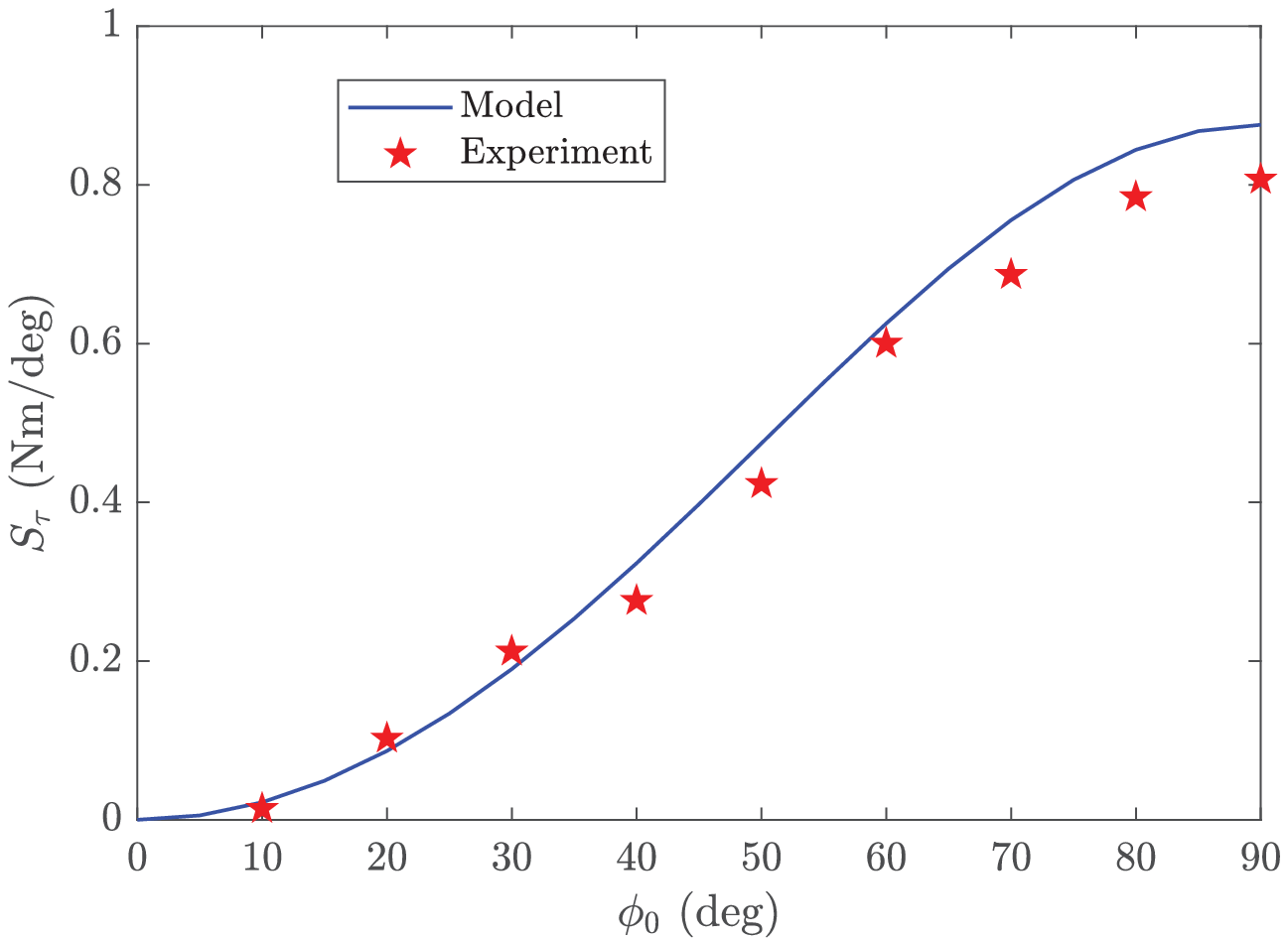}}
	\hspace{-2mm}
	\subfigure[]{
		\label{Fig_Steering_Sensitivity:c}
		\includegraphics[width=2.26in]{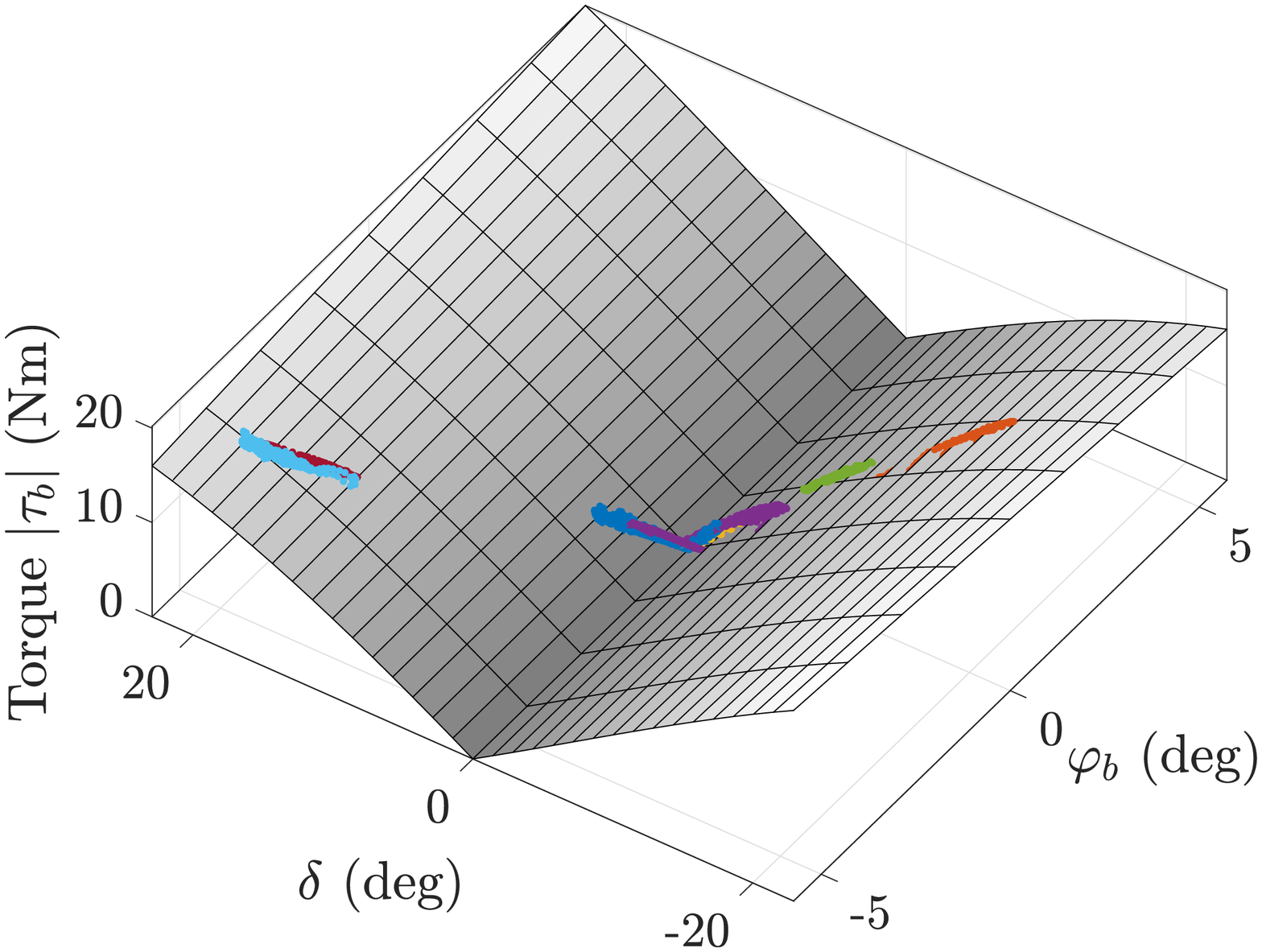}}
	\caption{Experimental results for the steering torque model. (a) Comparison results of the model prediction with the experiments for radius $r_{\phi_0}$. (b) Comparison of the steering sensitivity model prediction with the experiments. (c) Comparison of the balance torque model prediction with the experiments under steering angle increment $\delta$ and roll angle $\varphi_b$ with $\phi_0=90$ degs.}
	\label{Fig_Steering_Sensitivity}
	\vspace{-0mm}
\end{figure*}

Fig.~\ref{Fig_Bike_Photo} shows the prototype of the two-wheel steered bikebot with an onboard 6-DOF robotic manipulator (Jaco2 from Kinova Inc.). Fig.~\ref{Fig_Control_Board} illustrates the interconnection schematic of the embedded systems and actuators and sensors. Both the front and the rear wheels can be steered around $360$ degs by two stepper motors. A real-time low-level embedded system (Teensy 4.0 microcontroller) is used for the steering motor control, while the robotic manipulator is controlled by a powerful small-size computer (Intel NUC module) with robot operating system (ROS). One inertial measurement unit (IMU) (model 800 from Motion Sense Inc.) is mounted at the upper frame of the bikebot to measure the roll angle. The front and rear steering angles are measured by two encoders and the manipulator joint angles are obtained by the embedded encoders. The real-time bikebot steering control and data acquisition frequency was implemented at 100 Hz and the low-level manipulator velocity control was run at 1000 Hz.

\setcounter{figure}{3}
\begin{figure}[h!]
	\centering
	\includegraphics[width=3.2in]{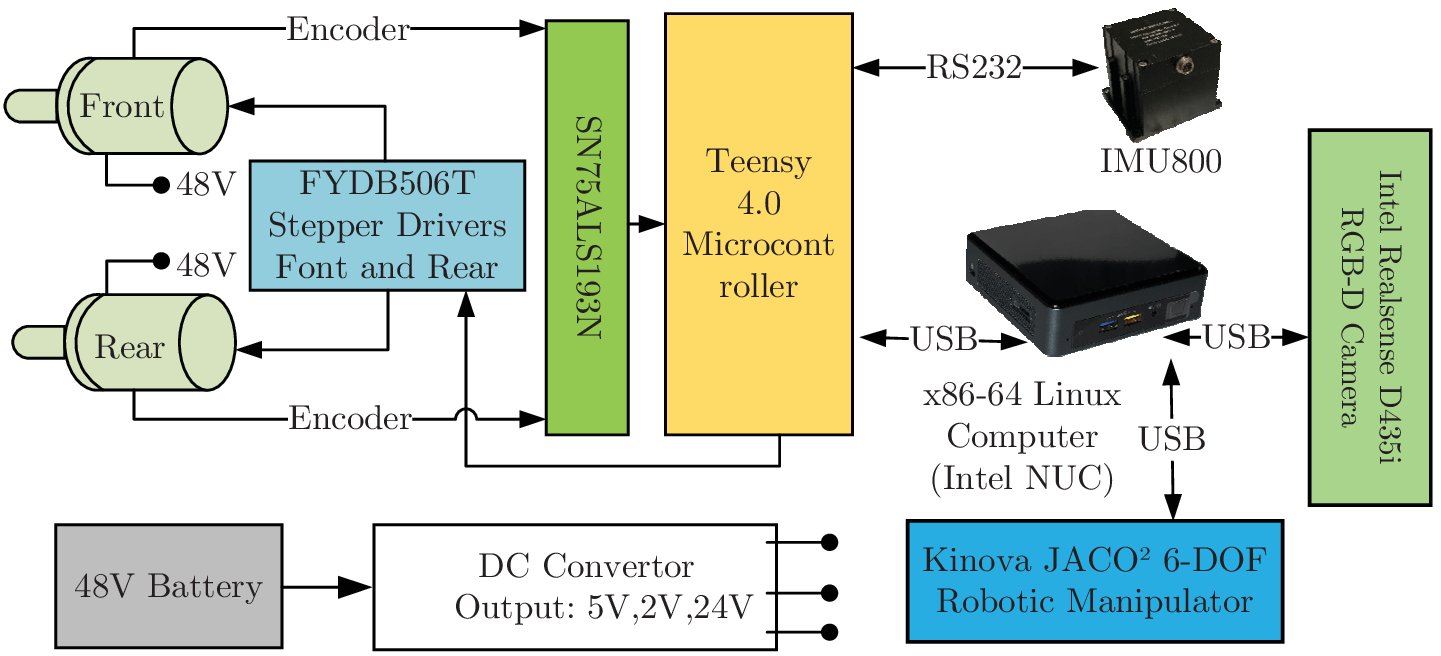}
	\caption{The schematic of the interconnection among sensors, actuators and embedded systems for the bikebot-manipulator system.}
	\label{Fig_Control_Board}
\end{figure}

\begin{figure}[h!]
	\centering
	\includegraphics[width=3in]{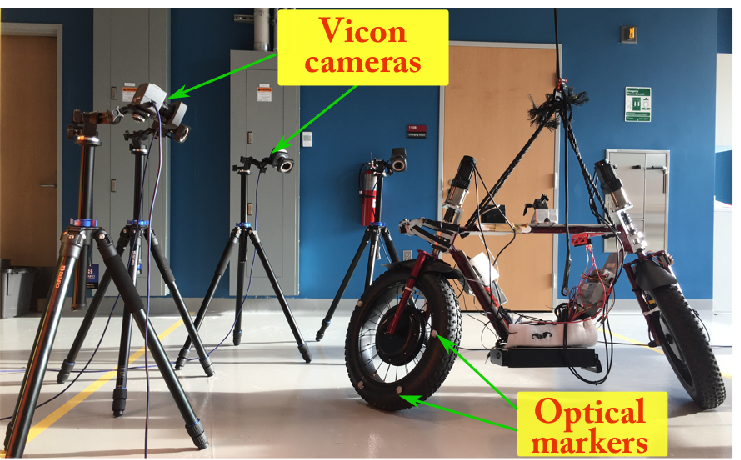}
	\caption{Experimental setup for validation of the steering mechanism and balance torque model.}
	\label{Fig_Steering}
	\vspace{-2mm}
\end{figure}

Table~\ref{Table_modelparameter} lists the model parameter values for the bikebot. The values of the DH parameters and mass moments of inertia of the manipulator links are listed in Table~\ref{Table_DH_Parameters}. The other physical parameters for each link can be found in~\cite{KinovaUG2019Jaco2}. To validate the steering mechanism and models, we also built and conducted experiments to measure the tire/ground contacts and movement. Fig.~\ref{Fig_Steering} shows the experimental setup for the steering models validation. A motion capture system (4 Vantage cameras from Vicon Ltd.) was used to measure the angle and contact points between the wheel surface and the ground at different initial steering angles $\phi_0$.

\setcounter{figure}{6}
\begin{figure*}[tb!]
	\hspace{-2mm}
	\subfigure[]{
		\label{Fig_Bike_Balance_Control:a}
		\includegraphics[width=2.3in]{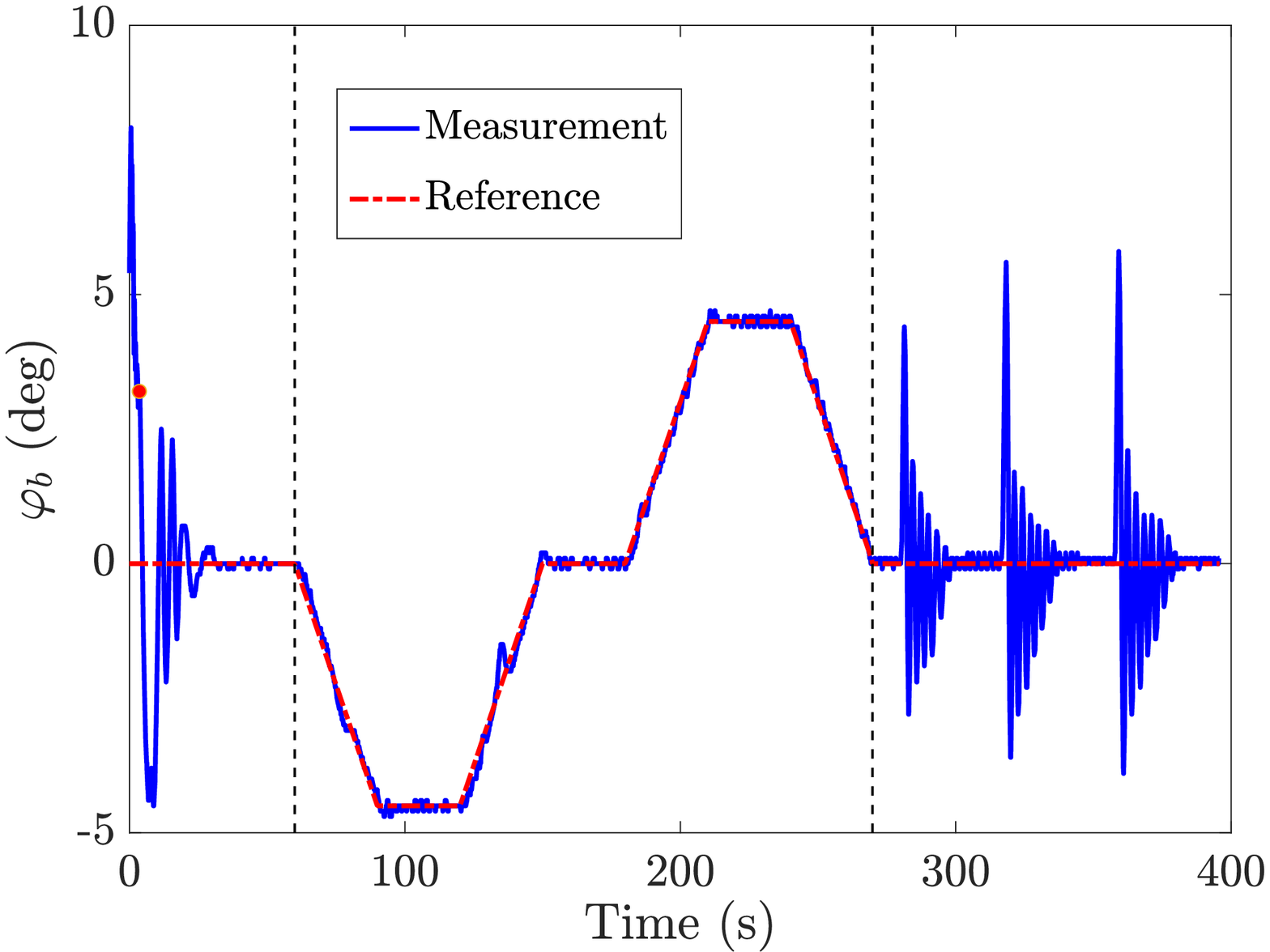}}
	\hspace{-1mm}
	\subfigure[]{
		\label{Fig_Bike_Balance_Control:b}
		\includegraphics[width=2.3in]{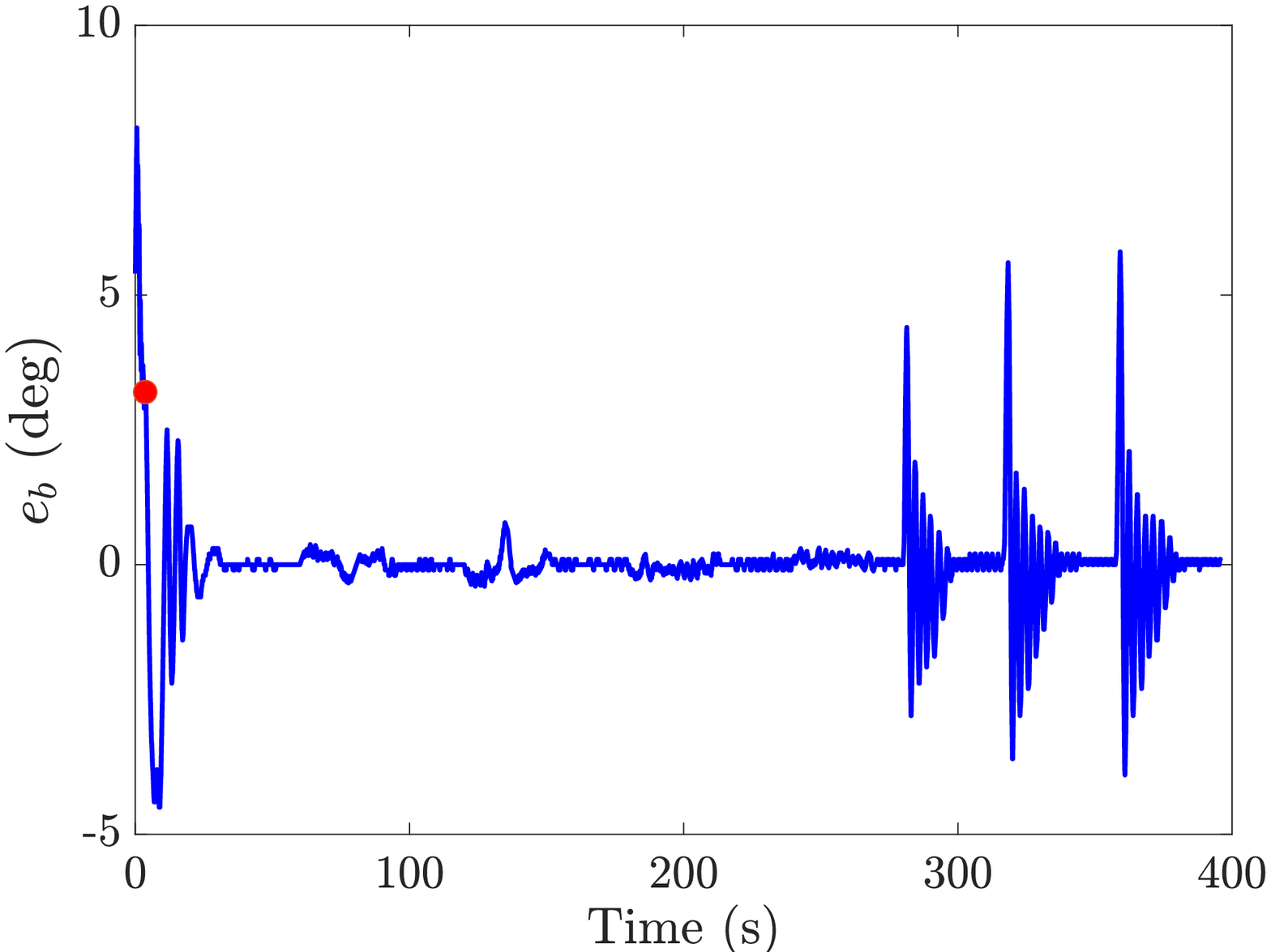}}
	\hspace{-1mm}
	\subfigure[]{
		\label{Fig_Bike_Balance_Control:c}
		\includegraphics[width=2.32in]{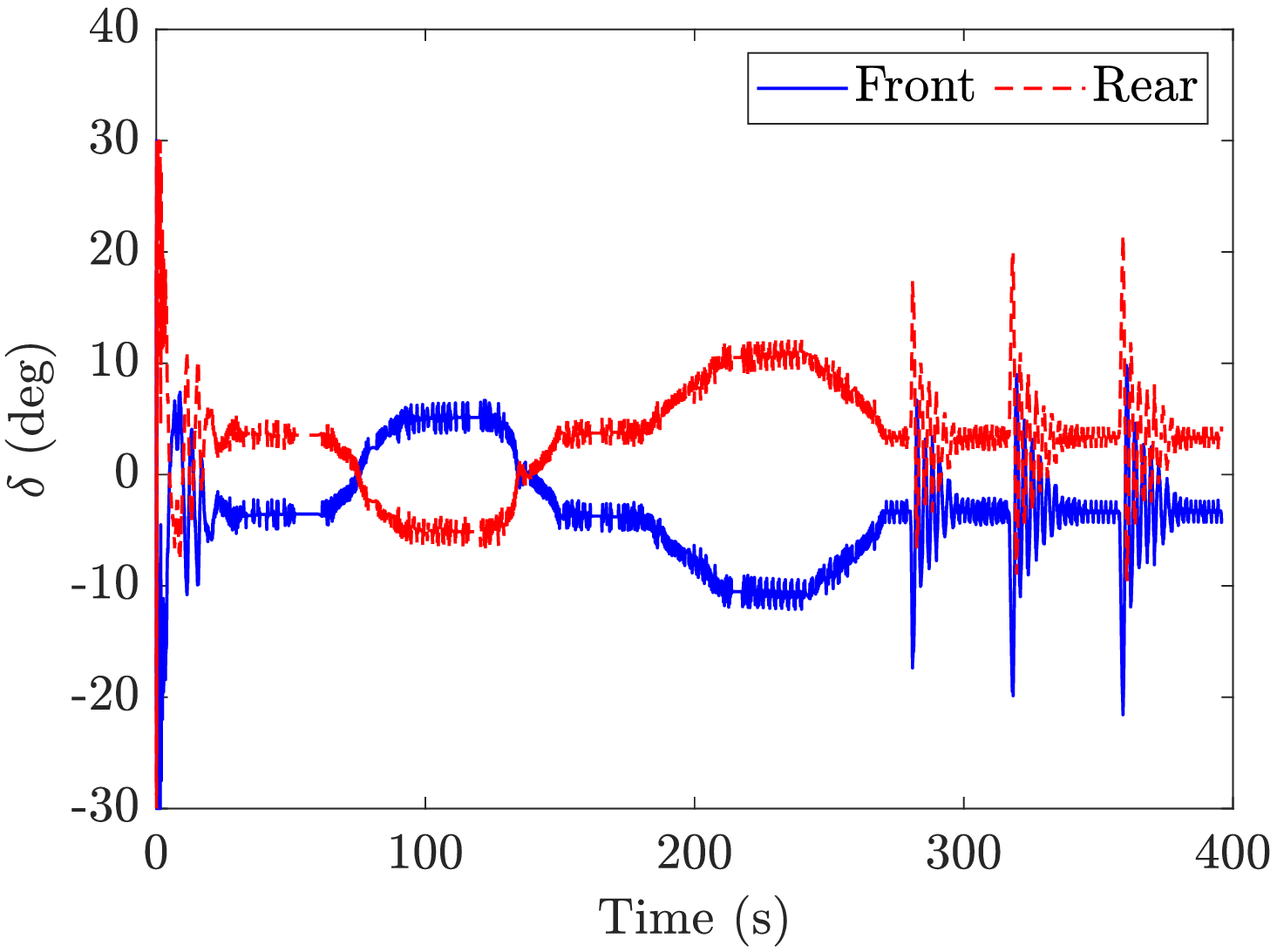}}
	\caption{Bikebot balance control experimental results. (a) Bikebot roll angle $\varphi_b$. (b) Bikebot roll angle error $e_b$. Steering angle increments. The red markers ``$\bullet$'' in (a) and (b) indicate that the initial angles.}
	\label{Fig_Bike_Balance_Control}
\end{figure*}

\renewcommand{\arraystretch}{1.3}
\setlength{\tabcolsep}{0.1in}
\begin{table}[h!]
	\centering
	\caption{Values for the model parameters of the bikebot platform}
	\label{Table_modelparameter}
	\vspace{-1mm}
	\begin{tabular}{|c|c|c|c|c|c|}
		\hline\hline
		$m_b$ (kg)& $J_b$ (kgm$^2$) & $h_G$ (m) & $l$ (m) & $\varepsilon$ (deg) & $R$ (m)  \\ \hline
		$46.9$ & $3.2$ & $0.53$ & $1.2$ & $20$ & $0.3$ \\	\hline\hline
	\end{tabular}
	\vspace{-2mm}
\end{table}

\renewcommand{\arraystretch}{1.3}
\setlength{\tabcolsep}{0.058in}
\begin{table}[t!]
	\vspace{-0mm}
	\centering
	\caption{DH parameter values and inertia parameters of the 6-DOF robotic manipulator}
	\vspace{-1mm}
	\begin{tabular}{|c|c|c|c|c|c|c|}
		\hline\hline
		Link & $\alpha_i$ (deg) & $a_i$ (m) & $d_i$ (m) & $m_i$ (kg) & $[I_{xx},I_{yy},I_{zz}]$ (kgm$^2$)\\
		\hline
		1 & $90$ & $0$ & $0.276 $ & $1$ & $[0.0022, 0.0006,0.0023]$\\
		\hline
		2 & $180$ &$ 0.41$ &0 & $1.5$ & $[0.0041,0.0255,0.0217]$\\
		\hline
		3 & $90$ &$0$ &$ -0.01$ & $0.8$ & $[0.0029,0.0027,0.0004]$\\
		\hline
		4 & $60$ &$0$ &$ -0.25$ & $0.3$ & $[0.7085, 0.7405, 0.1782]$\\
		\hline
		5 & $60$ & $0$ &$-0.009$ & $0.3$ & $[0.8275, 0.8520,0.1708]$\\
		\hline
		6 & $180$ &$0$ &$0.203$ & $0.6$ & $[0.0048,0.0048,0.0002]$\\
		\hline\hline
	\end{tabular}
	\label{Table_DH_Parameters}
	\vspace{-2mm}
\end{table}

\subsection{Experimental Results}
\renewcommand{\arraystretch}{1.3}
\setlength{\tabcolsep}{0.08in}
\begin{table*}
	\centering
	\caption{Collaborative end-effector pose control result. The unit for position is cm and for orientation is deg. The error mean and standard deviation values are calculated from $10$ s data of the pose holding phase.}
	\label{Table_Result}
	\vspace{-1mm}
	\begin{threeparttable}
		\begin{tabular}{|c|c|c|c|c|c|c|}
			\hline\hline
			\multirow{2}{*}{Pose} & \multirow{2}{*}{Desired $\bs \xi_{e}^k$ } & \multirow{2}{*}{BPIK-planned $\bs \xi_{e}( \mathcal{T}_{n+1}(\bs q^*_k))$}  & \multirow{2}{*}{Actual $\bs \xi_{e}^k$ }  & \multicolumn{2}{c|}{Errors} \\ \cline{5-6}
			&  &  &  & \multicolumn{1}{c|}{Position}  & \multicolumn{1}{c|}{Orientation}\\
			\hline
			1 & $[-13 \;-55 \;94\;-61\;52\;96]^T$ & $[-14\; -56\; 95 \;-64\;52\;98]^T$ & $[-14\;-56\;95\;-63\;51\;97]^T$ & $0.76\pm 0.02$  & $0.43\pm0.005$  \\
			\hline
			2 & $[-15\;-67\;109\;57\;31\;-57]^T$ & $[-15\;-65\;106\;58\;30\;-58]^T$ & $[-15\;-67\;109\;53\;30\;-58]^T$ & $0.34 \pm 0.01$ & $0.16\pm 0.017$\\
			\hline
			3 & $[-17\;-69\;106\;38\;55\;-37]^T$ & $[-16\;-68\;107\;37\;55\;-37]^T$ & $[-16\;-68\;107\;37\;55\;-37]^T$ & $0.72\pm 0.13$ & $0.40\pm0.08$\\
			\hline
			4 & $[-15\;-58\;114\;32\;48\;-21]^T$ & $[-15\;-61\;112\;39\;42\;-18]^T$ & $[-15\;-60\;112\;39\;42\;-18]^T$ & $0.34\pm 0.09$ & $0.15\pm0.046$ \\
			\hline\hline
		\end{tabular}
	\end{threeparttable}
\end{table*}

We first present the validation of the steering balance models. Figs.~\ref{Fig_Steering_Sensitivity:a} and~\ref{Fig_Steering_Sensitivity:b} show the values of the turning radius $r_{\phi_0}$ and the steering torque sensitivity $S_\tau$, respectively, as the static steering angle $\phi_0$ increases. The experiment data clearly confirm the model predictions. It is clear that when $\phi_0=90$ degs, the projected radius $r_{90}$ reaches the maximum value and the increasing trend is monotonic. The steering sensitivity $S_\tau$ also reaches its maximum point around $\phi_0=90$ degs with $S_\tau=0.87$ Nm/deg. At $\phi_0=0$, the projected radius $r_{0}$ and steering torque sensitivity $S_\tau$ are near zero. From this observation, an initial steering angle $\phi_0$ is chosen around $90$ degs for following stationary balance experiments. At $\phi_0=90$ degs, multiple stationary balancing experiments were conducted. Fig.~\ref{Fig_Steering_Sensitivity:c} shows the steering-induced balance torque $\tau_b$ at different roll angles $\varphi_b$ and increments $\delta$. Multiple experimental trials are plotted together with the steering torque model prediction from~(\ref{Eq_90_Steering_Torque}), i.e., the 3D surface as shown in the figure. The experimental data are scattered around the torque model prediction with small errors. These results validate the steering-induced balance torque model.

\setcounter{figure}{7}
\begin{figure}[htb!]
	\centering
	\includegraphics[width=3.2in]{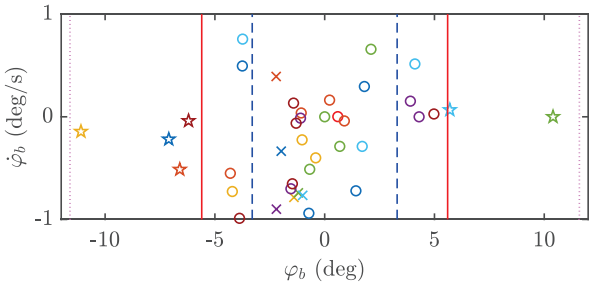}
	\caption{Verification of the recoverable roll angle region. The markers "$\times$", "$\circ$" and "$\star$" represent the successful balance trial of the bikebot by one-wheel steering, two-wheel steering, and collaboratively two-wheel steering with manipulation balancing strategies, respectively. The vertical lines indicate the estimated maximum angle boundaries $\varphi_b=\pm 3.4$, $\pm 5.6$, and $\pm 11.6$ degs.}
	\label{Fig_Balance_Capability}
	\vspace{-1mm}
\end{figure}

\setcounter{figure}{8}
\begin{figure*}[htb!]
	\centering
	\hspace{-3mm}
	\subfigure[]{
		\label{Fig_Bike_Arm_Move_3D:a}
		\includegraphics[width=2.5in]{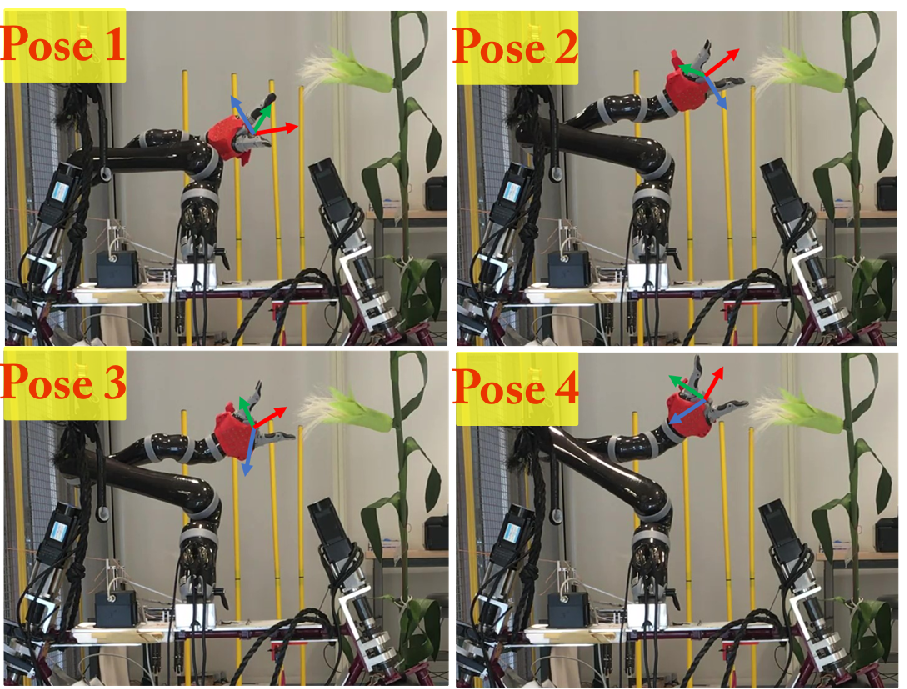}}
	\hspace{-2mm}
	\subfigure[]{
		\label{Fig_Bike_Arm_Move_3D:b}
		\includegraphics[width=1.7in]{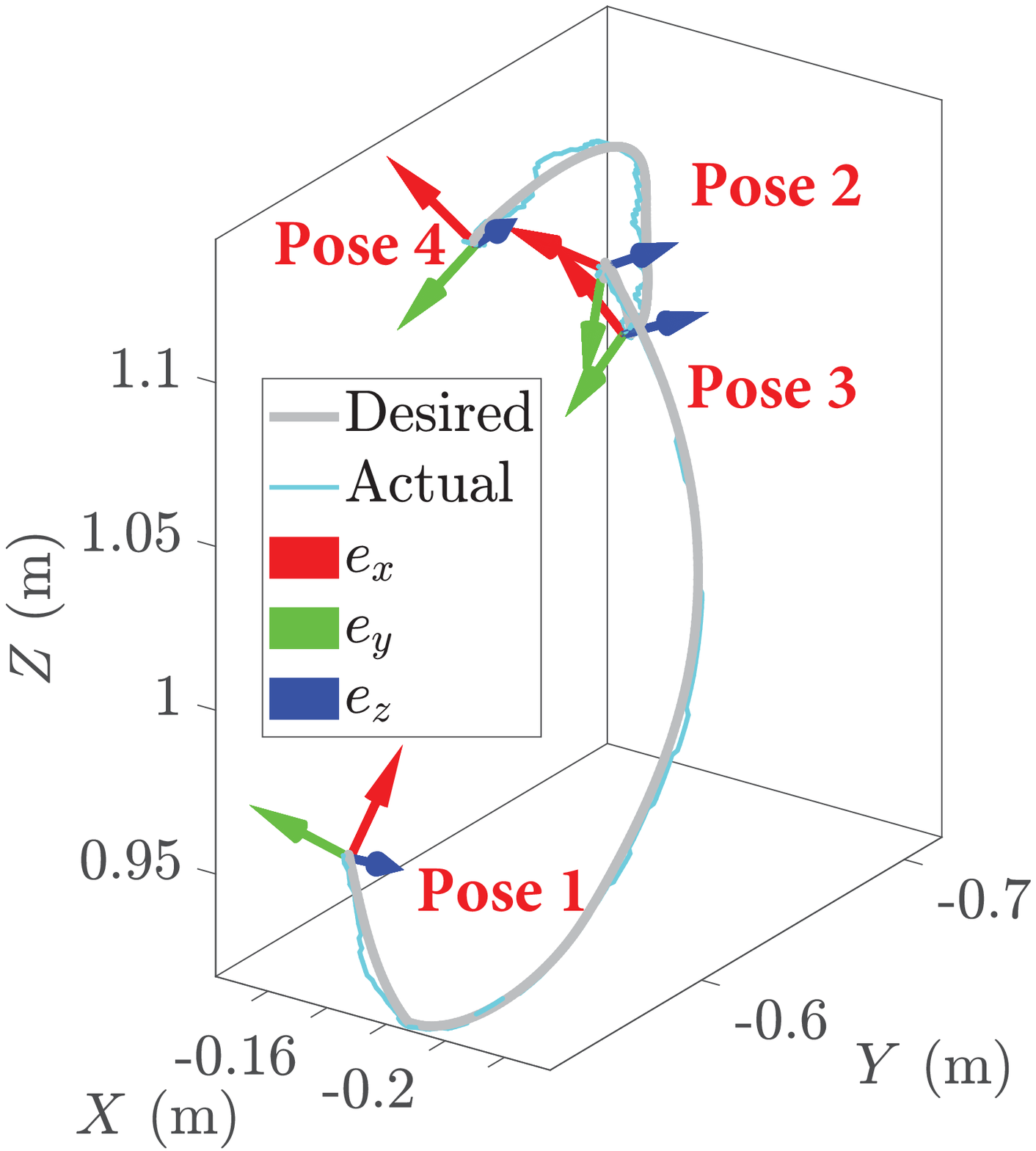}}
	\hspace{-2mm}
	\subfigure[]{
		\label{Fig_Error_Mean_Std}
		\includegraphics[width=2.8in]{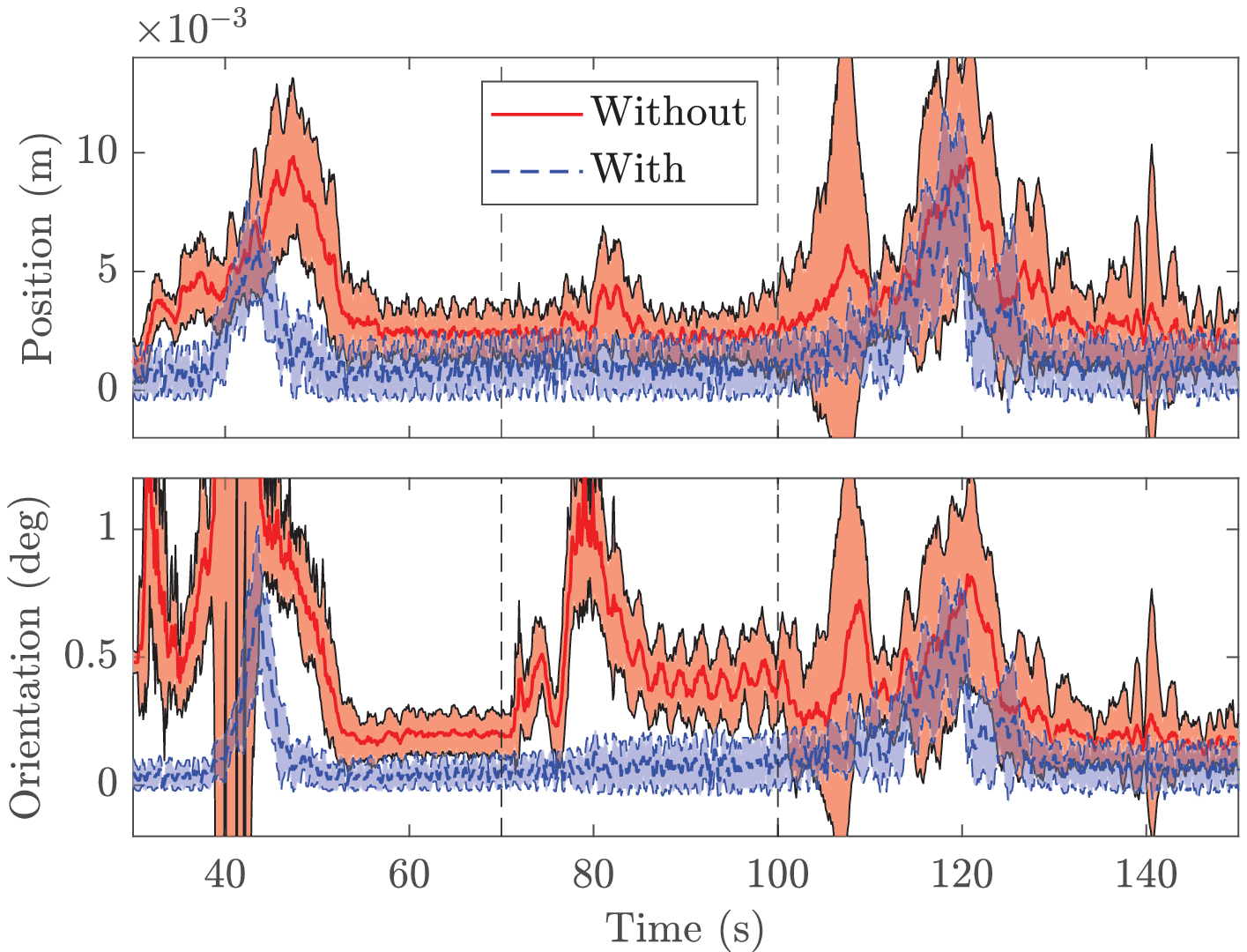}}
	\caption{(a) Snapshots of the manipulator with an inspected plant at four poses. (b) Pose transition and end-effector trajectory under the pose planning and control. The arrows (i.e., unit vectors $(\bs{e}_x,\bs{e}_y,\bs{e}_z)$) at each pose represent the actual orientation of the end-effector in $\mathcal{I}$. {(c) Pose error (mean with one standard deviation) in transition process (from Poses 1 to 4 as indicated by the vertical lines) from 15 experiment trials. Top: Position error; Bottom: Orientation error. The video of the experiment can be fount at: \href{https://youtu.be/jHQRNrrnPMc}{https://youtu.be/jHQRNrrnPMc}}.}
	\label{Fig_Bike_Arm_Move_3D}
	\vspace{-0mm}
\end{figure*}

Fig.~\ref{Fig_Bike_Balance_Control} shows the bikebot balance control results. The manipulator was removed from the bikebot in this experiment. The controller~(\ref{Eq_Balance_Control_PID}) was used with feedback gains $k_p=8.5$ and $k_d=2$. The entire trial is divided into three stages as separated by the vertical lines in Fig.~\ref{Fig_Bike_Balance_Control:a}. Fig.~\ref{Fig_Bike_Balance_Control:b} shows the roll angle tracking errors. Fig.~\ref{Fig_Bike_Balance_Control:c} illustrates the front and rear wheel steering angle increments. In the first stage, the initial roll angle was about $4$ degs and it was then regulated around zero. A chattering phenomenon was observed in the roll angle profile and this was due to the fact that the IMU angular measurement resolution was around $0.1$ deg, that is, the IMU measurement was discretized with a minimal resolution of $0.1$ deg. This oscillation also caused a similar chattering behavior in steering angle increments in Fig.~\ref{Fig_Bike_Balance_Control:c} since the roll angle measurement was used in steering control. In the second stage starting at around $t=60$ s, the bikebot was commanded to move back and forth around the zero with the maximum roll angles around $4.5$ degs. The change of the reference roll angle was slow to meet the quasi-static movement. The tracking error approached to zero. In the third stage starting around at $t=270$~s, multiple disturbances were applied by manually pushing upper frame of the bikebot. The roll angle errors caused by the disturbances reached $6$ degs and the steering actuation compensated for the disturbances. These results demonstrate the steering balance control performance. Since the design enforced symmetrical steering commands, the actual front and rear steering angles' responses showed highly similar behaviors.

Using the BEM and model parameters, we estimate the maximum stationary balance roll angles. With one-wheel steering control, the maximum balanced roll angle is around $3.4$ degs; with two-wheel steering control, around $5.6$ degs; and additionally, if the manipulator is used to help balance collaboratively, it increases to $11.6$ degs. To validate these estimates, we conducted multiple balance control experiments. One experimental trial was considered successful if the system was kept balance for a time period over $50$ s. Fig.~\ref{Fig_Balance_Capability} shows the successful trials in the $\dot{\varphi}_b$-$\varphi_b$ plane. Each marker in the figure represents the state at which the bikebot successively started to balance under steering control, which was confirmed by comparing the model predictions from~(\ref{Eq_90_Steering_Torque}) and dynamics~(\ref{Eq_Bike_Roll}). The experiments, which are in agreement with the model prediction, validate the model analysis and demonstrate the balance capability under various bikebot balancing strategies.

\setcounter{figure}{9}
\begin{figure*}[thb!]
	\centering
	\subfigure[]{
		\label{Fig_Bike_Arm_Move:a}
		\includegraphics[width=2.32in]{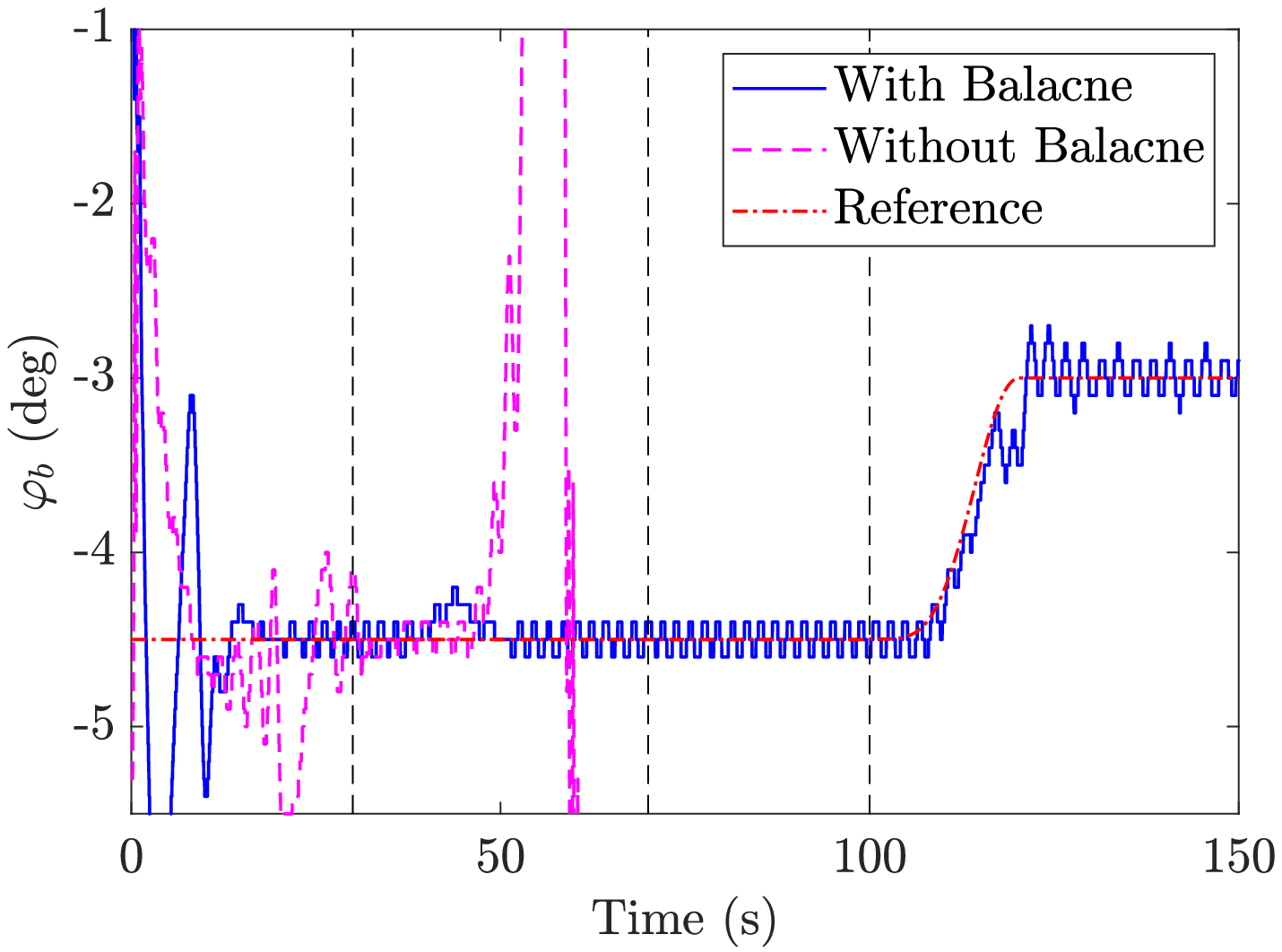}}
	\hspace{-1mm}
	\subfigure[]{
		\label{Fig_Bike_Arm_Move:b}
		\includegraphics[width=2.3in]{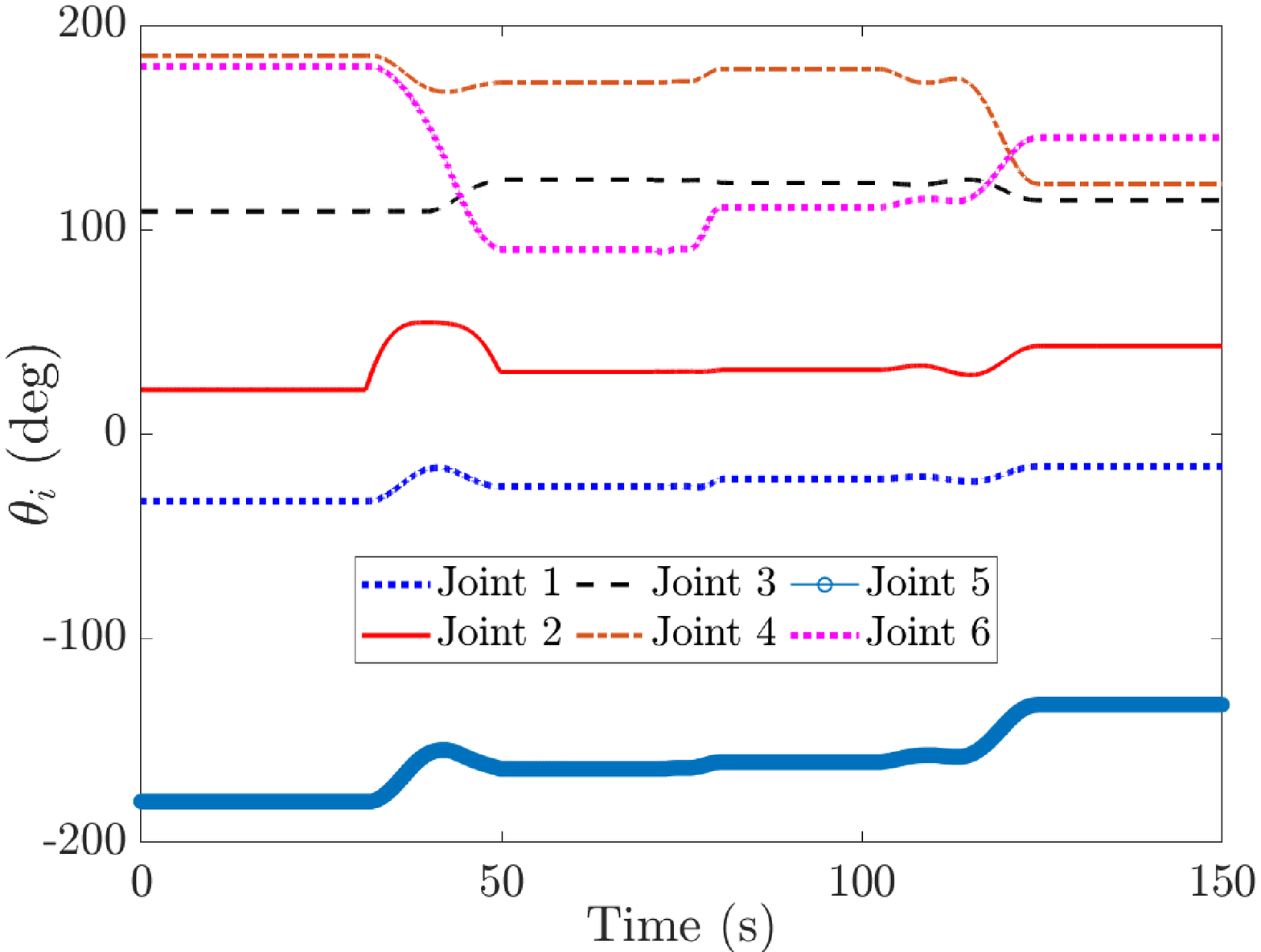}}
	\hspace{-1mm}
	\subfigure[]{
		\label{Fig_Bike_Arm_Move:c}
		\includegraphics[width=2.26in]{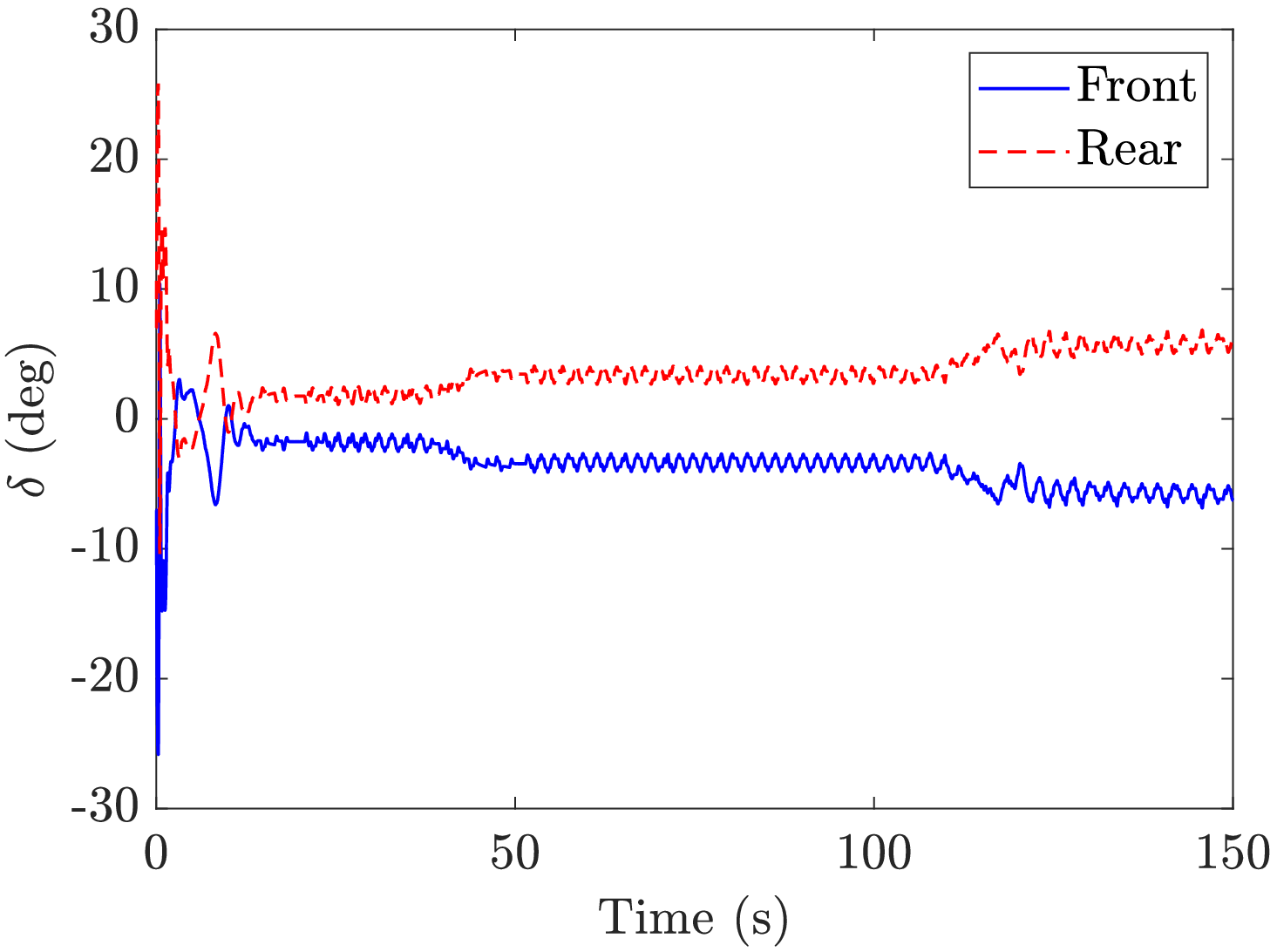}}
	\hspace{-1mm}
	\subfigure[]{
		\label{Fig_Bike_Arm_Move:d}
		\includegraphics[width=2.21in]{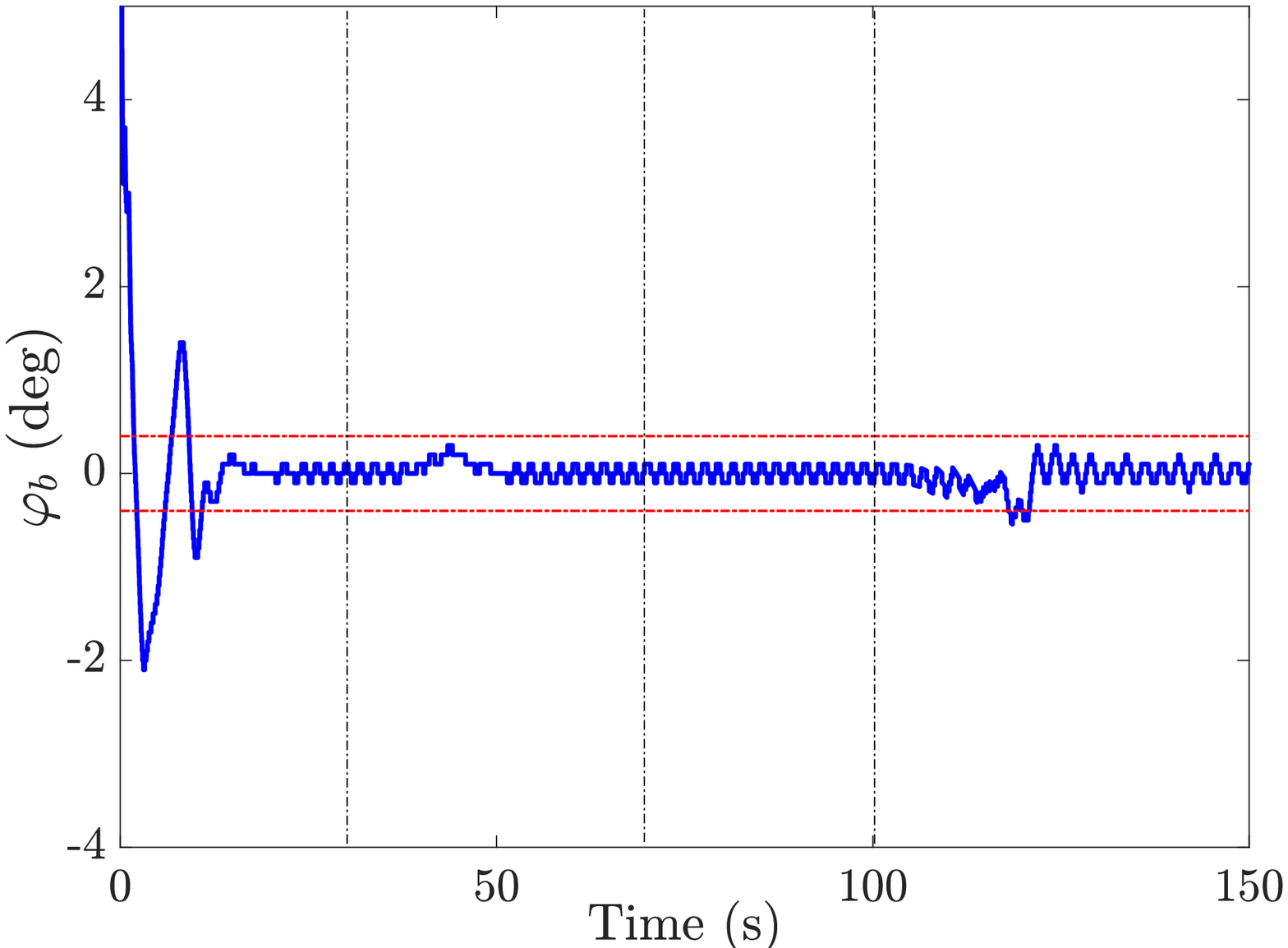}}
	\hspace{-1mm}
	\subfigure[]{
		\label{Fig_Bike_Arm_Move:e}
		\includegraphics[width=2.27in]{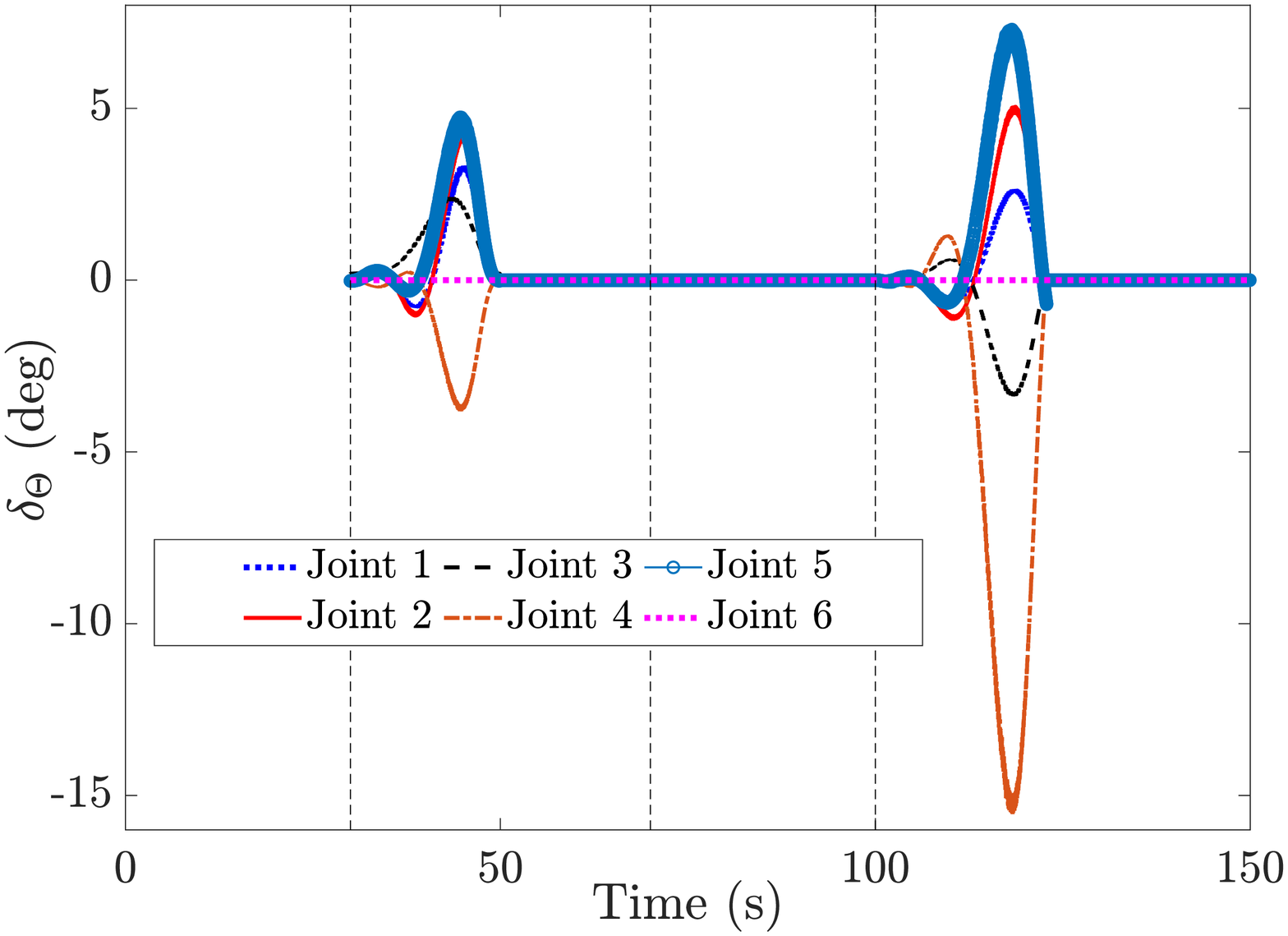}}
	\vspace{-2mm}
	\caption{Large roll angle balance control of the mobile manipulation system. (a) {Roll angle}, (b) robotic arm joint angles, and (c) steering angle increments. The vertical lines in (a) divides the entire process into four phases for each pose. (d) Roll angle error. (e) Online robotic arm trajectory correction in pose transition phase (the difference between off-line planing results and the actual angle).}
	\label{Fig_Bike_Arm_Move}
	\vspace{-2mm}
\end{figure*}

We now present a plant inspection example for the end-effector to continuously go through and stop momentarily at four poses (i.e., $N_\xi=4$). This represents the end-effector movement during a plant scanning and inspection task~\cite{EdmondsCASE2019}. Fig.~\ref{Fig_Bike_Arm_Move_3D:a} shows the snapshots of the end-effector at the four poses. The major movement of the end-effector (with a mounted camera) was along the $z$-axis in $\mathcal{I}$ and orientation always pointed towards to the stalk of a fake corn plant. The end-effector moved from one pose to another in sequence and stopped for about 15 s at each pose. Fig.~\ref{Fig_Bike_Arm_Move_3D:b} shows the 3D trajectories with the four poses. The planning and control parameters used in experiments include: $\lambda_1=10$, $\lambda_2=1$, $\lambda_3=5$, $\lambda_4=1.5$, $\bs W_1=\diag(10,5,5,5,1,1,1)$,  $\bs W_2=\diag(1,1,1,1,1,1)$, $\kappa=5$, $\epsilon=0.1$, $\varepsilon_b=0.4$ degs, $\dot{q}_{\max}=36$ deg/s, $\ddot{q}_{\max}=120$ deg/s$^2$, $\bs{\tau}^{\max}_\theta=[10\; 15\; 10\; 5\; 5\; 5]^T$ Nm, $\delta_{\max}=15$ degs, $\dot{\delta}_{\max}=20$ deg/s, and the degree of the B\'{e}zier polynomial $N=7$. For off-line planning implementation, the number of data points was chosen as $N_s=50$ in each dimension of $\bm q(t)$ and the SQP method (via \emph{fmincon} function) in Matlab was used for solving~(\ref{Eq_Trajectory_Plan}). The obtained  B\'{e}zier polynomial trajectory was then sampled at 100 Hz for real time control. The off-line planner computed the trajectory for the next pose transition when conducting the real-time motion control of the current pose movement. By doing so, the proposed planner was capable of obtaining the trajectory with fast computational time. Table~\ref{Table_Result} lists the desired end-effector poses, the poses planned by the BPIK and the actual poses under the controller. From the table, the position errors are within 8 mm and orientation errors within $0.45$ degs at these four poses.

Fig.~\ref{Fig_Bike_Arm_Move} shows the detailed experimental results. Figs.~\ref{Fig_Bike_Arm_Move:a} and~\ref{Fig_Bike_Arm_Move:b} show the bikebot roll angle $\varphi_b$ and the six joint angles $\bs{\Theta}$ of the manipulator, respectively. Since $\varphi_b$ and the first three joint angles ($\theta_1$-$\theta_3$) played a major role to balance the entire system, their reference trajectories were designed to avoid large variations as shown in the figures. Poses 2 and 3 were searched in the local workspace $\mathcal{X}_{\varphi_{b1}^0}(\bs{\Theta})$ of Pose 1, and Pose 4 is searched in the workspace $\mathcal X(\bs{q})$. The bikebot roll angle change was approximately around $1.5$ degs. At $t=0$~s, the manipulator was at the desired balance configuration as Pose 1. Around $t=30$~s, the manipulator started moving to Pose 2. Small disturbances were introduced at around $t=40$~s, causing about a $0.4$-deg roll angle error. The velocity correction control was applied to compensate for the roll angle error; see Fig.~\ref{Fig_Bike_Arm_Move:e}. No obvious roll angle error was observed during the transition from Poses 2 to 3 (except around $0.1$ degs oscillation). The bikebot platform was required to move in the transition from Poses 3 to 4. Around $t=110$~s, a large roll angle change was commanded by the steering actuation and the velocity correction control was needed; see Fig.~\ref{Fig_Bike_Arm_Move:e}.

\subsection{Discussion}
To further examine the performance, we conducted additional comparison experiments. Fig.~\ref{Fig_Bike_Arm_Move:a} shows the bikebot roll angle when the bikebot balance priority is not considered in the trajectory planning. Clearly, the entire system lost balance in the pose transition phase at $t=50s$. The result confirms the challenge in manipulator pose movement control. We further conducted and repeated the 4-pose control experiment 15 times. Fig.~\ref{Fig_Error_Mean_Std} summarizes the end-effector pose errors statistics (i.e., mean and one-standard deviation) with and without online velocity correction under the proposed trajectory planning method as modeling errors and system uncertainties might exist. With the online modification the end-effector position errors are less than $5$ mm and the orientation errors within $0.3$ degs. Relative large errors happened around $t=42$ and $120$ s with about $10$ mm and $0.7$ degs, respectively. This is consistent with the previous results. The position errors are at the same level of the manipulator hardware performance limits ($3.7$ mm) that are provided by the vendor and the orientation errors are much less that level ($2.1$ degs). Without the velocity correction, both the position and orientation tracking errors are larger than these with correction. The results demonstrate the successful balance and pose control performance by the design. We also conducted computational time comparison between the proposed trajectory planning algorithm and the DP method and the result is shown in Table~\ref{Table_ComputationCost}. The numerical results confirmed that the computational cost of the DP method was over 200 times higher than that of the proposed algorithm to solve the optimization problem in~(\ref{Eq_Trajectory_Plan}).

\renewcommand{\arraystretch}{1.3}
\setlength{\tabcolsep}{0.1in}
\begin{table}[tb!]
	\centering
	\caption{Computational cost comparison for the proposed and the DP algorithms}
	\vspace{2mm}
	\begin{tabular}{|c|c|c|c|c|}
		\hline\hline $N_s$  & $50$ & $100$ & $200$ & $500$ \\
		\hline Proposed algorithm (sec) &$9.69$ & $14.94$ & $31.58$ & $21.46$\\
		\hline DP method (hour) & $0.14$ & $1.04$ & $3.02$ & $\ge 12$ \\
		\hline\hline
	\end{tabular}
	\label{Table_ComputationCost}
    \vspace{-3mm}
\end{table}

{\color{red}} Although demonstrating successful results, the current work have several limitations for further improvement. We only studied coordinated control of the manipulator on stationary bikebot and it would be desirable to extend to moving platform case. Second, the steering control in this work does not include the dynamic effects of the steering mechanism. The control performance might be improved with incorporating dynamic steering effect. The trajectory of the bikebot roll angle and the manipulator joint angles was planned off-line and online planning is desirable for applications with dynamic obstacle avoidance. Finally, the proposed method is built on the accurate model of the system and it is desirable to extend to handle model parameter uncertainties in complex, dynamic environment. One possible approach is to use machine learning-based methods. For example, as discussed in~\cite{HanRAL2021}, the robot dynamics might be approximated and estimated using a Gaussian process learning approach and a learning-based motion control can be then designed.

\section{Conclusion}
\label{Section_Conclusion}

We presented a coordinated balance and pose control for a stationary mobile manipulation using a two-wheel steered bikebot. The mobile platform is inherently unstable and the dynamics of the platform and the manipulator are strongly coupled. We first presented a two-wheel steering model and identified the use of $\phi_0=90$ degs as the most beneficial steering angle for stationary balance. A balance equilibrium manifold was extended to the mobile manipulation for coordinated motion control. Built on the BEM, a balance-priority design was presented to solve the optimal joint angles for the bikebot and the manipulator. Coordinated balance and pose control was achieved by enforcing the entire system moving on the BEM with online manipulator velocity correction control. We conducted extensive experiments and the results demonstrated the performance of the balance and pose control for a plant inspection application. 

\bibliographystyle{IEEEtran}
\bibliography{YiRef,Reference}

\end{document}